%% file: main.tex
\documentclass[runningheads]{llncs}

\usepackage{eccv}

\usepackage{eccvabbrv}

\usepackage{graphicx}
\usepackage{booktabs}

\usepackage[accsupp]{axessibility}  

\input{preamble}

\usepackage{hyperref}

\usepackage{orcidlink}

\begin{document}

\title{V-Co: A Closer Look at Visual Representation Alignment via Co-Denoising} 

\titlerunning{V-Co}


\author{
Han Lin\inst{1}\orcidlink{0000-0002-1303-8636}  \and
Xichen Pan\inst{2} \and
Zun Wang\inst{1}\orcidlink{0009-0005-9502-050X} \and
Yue Zhang\inst{1}\orcidlink{0000-0003-2153-6536} \and
Chu Wang\inst{3} \and \\
Jaemin Cho\inst{4,5}\orcidlink{0000-0002-1558-6169} \and
Mohit Bansal\inst{1}\orcidlink{0009-0009-2965-5354}}

\authorrunning{H. Lin et al.}

\institute{University of North Carolina at Chapel Hill, USA \and New York University, USA \and Meta, USA \and AI2, USA \and Johns Hopkins University, USA\\
\url{https://github.com/HL-hanlin/V-Co}}

\maketitle

\begin{abstract}
Pixel-space diffusion has recently re-emerged as a strong alternative to latent diffusion, enabling high-quality generation without pretrained autoencoders. However, standard pixel-space diffusion models receive relatively weak semantic supervision and are not explicitly designed to capture high-level visual structure. Recent representation-alignment methods (\eg{}, REPA) suggest that pretrained visual features can substantially improve diffusion training, and \emph{visual co-denoising} has emerged as a promising direction for incorporating such features into the generative process. However, existing co-denoising approaches often entangle multiple design choices, making it unclear which are truly essential. We therefore present \textbf{V-Co}, a systematic study of visual co-denoising in a unified JiT-based framework. This controlled setting allows us to isolate the ingredients that make visual co-denoising effective. Our study reveals two main ingredients. First, co-denoising benefits from preserving feature-specific computation while enabling flexible cross-stream interaction, which leads to a fully dual-stream architecture together with a structurally defined unconditional prediction for classifier-free guidance. Second, it requires both stronger semantic supervision and proper cross-stream calibration, which we realize through a perceptual-drifting hybrid loss and RMS-based feature rescaling. Together, these findings yield a simple recipe for visual co-denoising. Experiments on ImageNet-256 show that, at comparable model sizes, V-Co outperforms the underlying pixel-space diffusion baseline and strong prior pixel-diffusion methods while using fewer training epochs, offering practical guidance for future representation-aligned generative models.
\end{abstract}

\input{sec/1_intro}
\input{sec/2_related_work}

\input{sec/3_method}
\input{sec/4_experiments}

\input{sec/5_conclusion}

%
%
\clearpage
\bibliographystyle{splncs04}
\bibliography{main}

\appendix
\onecolumn
\input{sec/99_appendix}

\end{document}

%% file: preamble.tex
\usepackage[table]{xcolor}
\definecolor{lightblue}{rgb}{0.90, 0.95, 0.98}
\usepackage{xcolor}
\definecolor{customblue}{HTML}{C3D7F3}
\definecolor{softgreen}{HTML}{D9E7D5}
\definecolor{softpurple}{HTML}{BDAEE5}

\usepackage{multicol}
\usepackage{titletoc}
\usepackage{amsmath}  
\usepackage{bm}       
\usepackage{wrapfig}
\usepackage{xspace}
\usepackage{algorithm, algpseudocode}
\usepackage{minted}
\usepackage{multirow}
\usepackage{booktabs}
\usepackage{makecell}
\usepackage{boldline}
\usepackage{array,multirow,graphicx}
\usepackage{enumitem}
\usepackage[dvipsnames]{xcolor}
\usepackage{amsmath, amssymb, amsfonts}

\newcommand{\TODO}[1]{{\color{black}#1}}

\definecolor{lightgreen}{rgb}{0.565, 0.933, 0.565}
\definecolor{lightpurple}{rgb}{0.784, 0.635, 0.784}

\usepackage[most]{tcolorbox}
\usepackage{xcolor}

\definecolor{myborder}{RGB}{0,0,0}
\definecolor{mygray}{RGB}{242,242,242}

\newtcolorbox{questionbox}{
  colback=mygray,
  colframe=myborder,
  boxrule=0.8pt,
  arc=8pt,
  left=6pt,
  right=6pt,
  top=2pt,
  bottom=2pt,
  enhanced,
  drop shadow={black!18!white},
}

%% file: sec/1_intro.tex
\section{Introduction}
\label{sec:intro}

Diffusion models~\cite{esser2024scaling, peebles2023scalable, ma2024sit} have achieved remarkable success in image generation. While much recent progress has been driven by latent diffusion models~\cite{Rombach_2022_CVPR} (LDMs), which denoise in compressed autoencoder spaces~\cite{kingma2013auto, Rombach_2022_CVPR}, an increasingly compelling alternative is pixel-space diffusion with scalable Transformer-based denoisers~\cite{li2025back, lu2026one, chen2025dip, yu2025pixeldit, ma2026pixelgen}. Recent systems such as JiT~\cite{li2025back} show that direct pixel-space denoising can be competitive while avoiding autoencoder-induced biases and bottlenecks. However, pixel-level denoising objectives are not explicitly designed to enforce high-level semantic structure, making semantic representation learning and composition less sample-efficient.

\begin{wrapfigure}[22]{r}{0.55\linewidth}
  \centering
  \includegraphics[width=\linewidth]{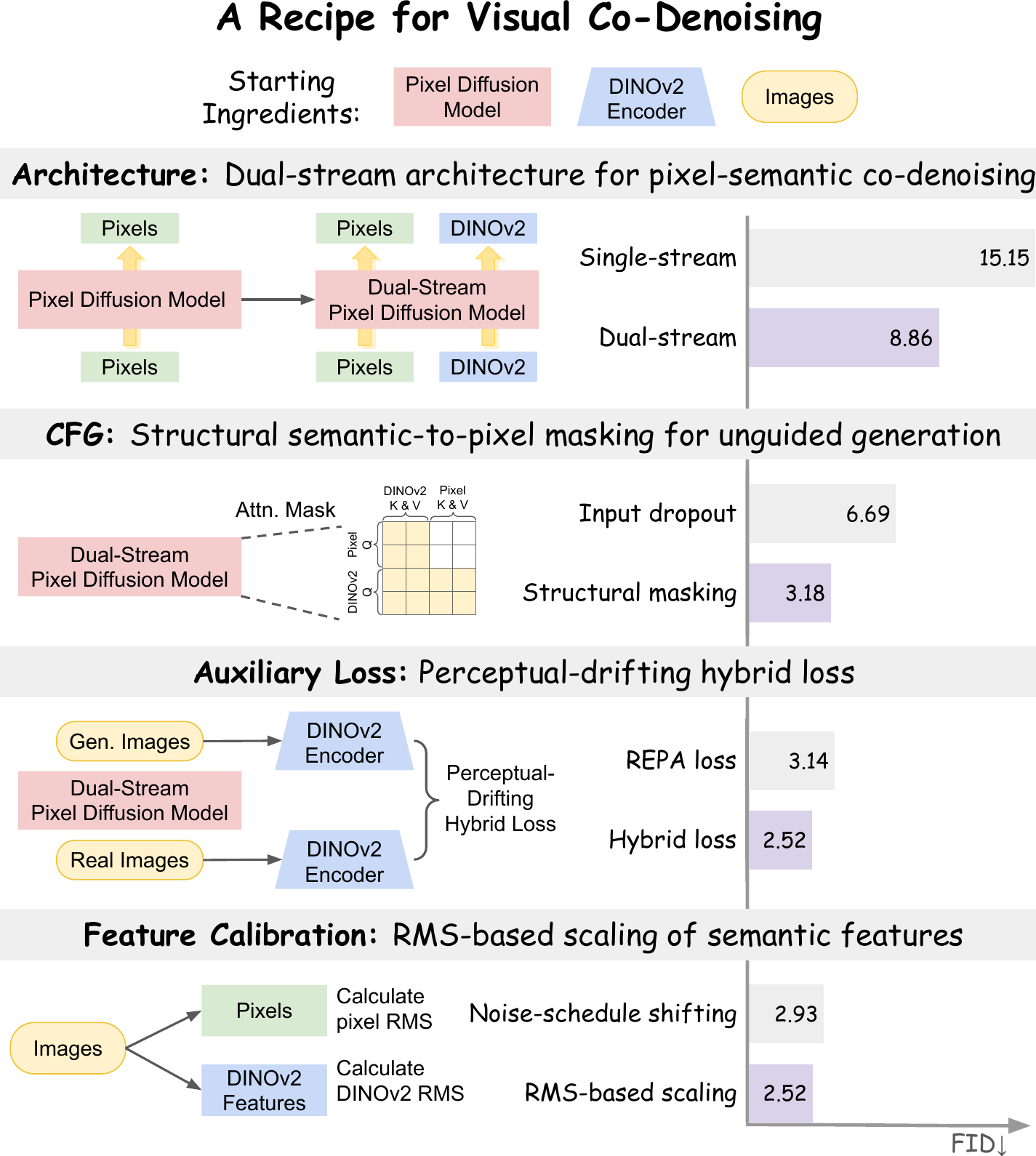}
  \caption{{\textbf{Overview of V-Co.}}
  }
  \label{fig:overview}
\end{wrapfigure}
In parallel, a growing body of work has explored how to inject external visual knowledge from strong pretrained encoders into diffusion training. One line of research adds \emph{representation-alignment losses} that encourage diffusion features to match pretrained visual representations~\cite{zhang2025videorepa, wang2026vae, tian2025u, wang2025repa, zheng2025flare, singh2025irepa}. Another performs denoising directly in a \emph{representation latent space}, rather than in pixel or VAE latent space~\cite{tong2026scaling, hu2025meanflow, chang2026dino, zheng2025diffusion}. A third line of work explores \emph{joint generation or co-denoising} architectures, in which image latents are generated together with semantic features or other modalities so that the streams can exchange information throughout the denoising trajectory~\cite{baade2026latent, han2025tv2tv, chefervideojam, kouzelisboosting, zhong2025flowvla, dangsvimo, bi2025motus, huang2025unityvideo, wu2025does, yang2025echomotion, dang2025syncmv4d}. 
Among these directions, visual co-denoising provides a deeper form of integration by incorporating pretrained semantic representations directly into the denoising dynamics, rather than using them only as supervision or as an alternative latent space. However, existing co-denoising systems typically entangle multiple design choices (including architecture, guidance strategy, auxiliary supervision, and feature calibration), making it unclear which ingredients are most important and how to combine them into a robust and scalable recipe.

In this paper, we study visual co-denoising as a mechanism for visual representation alignment. Rather than treating co-denoising as a fixed end-to-end design, we ask what makes it effective. To this end, we build a unified pixel-space testbed on top of JiT~\cite{li2025back}, where an image stream is jointly denoised with patch-level semantic features from a frozen pretrained visual encoder (\eg{}, DINOv2~\cite{oquab2023dinov2}). Within this controlled framework, we investigate four key questions: \textbf{(i)} what architecture best balances feature-specific processing and cross-stream interaction; \textbf{(ii)} how to define the unconditional branch for classifier-free guidance; \textbf{(iii)} which auxiliary objectives provide the most effective complementary supervision; and \textbf{(iv)} how to calibrate semantic features relative to pixels during diffusion training. Our goal is not only to improve performance, but also to distill general principles for effective co-denoising.

Based on this study, we derive a simple yet effective \textbf{V}isual \textbf{Co}-Denoising (\textbf{V-Co}) recipe, illustrated in~\cref{fig:overview}.
First, from the perspective of {model architecture}, we show that effective visual co-denoising requires preserving feature-specific computation while enabling flexible cross-stream interaction. Among a broad range of shared-backbone and fusion-based variants, a \emph{fully dual-stream} JiT consistently delivers the strongest performance (\cref{subsec:model_architecture}). 
Second, for classifier-free guidance (CFG), we introduce a novel \emph{structural masking} formulation, where unconditional prediction is defined by explicitly masking the semantic-to-pixel pathway rather than by input-level corruption alone. This simple design proves substantially more effective than standard dropout-based alternatives in co-denoising (\cref{subsec:model_architecture_unconditional_generation}). 
Third, we observe that instance-level semantic alignment and distribution-level regularization play complementary roles, and leverage this insight to propose a novel \emph{perceptual-drifting hybrid loss} that combines both within a unified objective, yielding the best generation quality in our study (\cref{subsec:auxiliary_loss}). 
Finally, we show that RMS-based feature rescaling admits an equivalent interpretation as a semantic-stream noise-schedule shift via signal-to-noise ratio (SNR) matching, providing a simple and principled calibration rule for cross-stream co-denoising (\cref{subsec:rms_scaling_time_shift}). 
Together, these contributions transform visual co-denoising into a concrete recipe for visual representation alignment.

Empirically, V-Co yields strong gains on ImageNet-256 under the standard JiT~\cite{li2025back} training protocol. Starting from a pixel-space JiT-B/16 backbone, our progressively improved recipe substantially outperforms both the original JiT baseline and prior co-denoising baselines (see \Cref{table:imagenet_256_comparison}), and achieves strong guided generation quality. 
Notably, V-Co-B/16 with only 260M parameters, matches JiT-L/16 with 459M parameters (FID 2.35 \vs{} 2.36). V-Co-L/16 and V-Co-H/16, trained for 500 and 300 epochs respectively, outperform JiT-G/16 with 2B parameters (FID 1.71 \vs{} 1.82) and other strong pixel-diffusion methods.

In summary, our contributions are three-fold:
\begin{itemize}[leftmargin=1.2em]
    \item We present a principled study of visual representation alignment via co-denoising (\textbf{V-Co}) in pixel-space diffusion, systematically isolating the effects of architecture, CFG design, auxiliary objectives, and feature calibration.
    \item We introduce an effective recipe for visual co-denoising with two key innovations: \emph{structural masking} for unconditional CFG prediction and a \emph{perceptual-drifting hybrid loss} that combines instance-level alignment with distribution-level regularization. Our study further identifies a fully dual-stream architecture and RMS-based feature calibration as the preferred design choices.
    \item We show that these designs yield strong improvements on ImageNet-256~\cite{deng2009imagenet}, outperforming the underlying pixel-space diffusion baseline (\ie{}, JiT~\cite{li2025back}) as well as prior pixel-space diffusion methods.
\end{itemize}

%% file: sec/2_related_work.tex
\section{Related Work}
\label{sec:related_work}

\textbf{Pixel-space diffusion generation.}
Recent work has shown that, with suitable architectural and optimization choices, diffusion models trained directly in pixel space can approach latent diffusion performance~\cite{Rombach_2022_CVPR}. JiT~\cite{li2025back} demonstrates that competitive pixel-space generation is possible with a minimalist Transformer design, while Simple Diffusion~\cite{hoogeboom2023simple}, PixelDiT~\cite{yu2025pixeldit}, and HDiT~\cite{crowson2024scalable} improve training and scalability. Other methods add stronger inductive biases, such as decomposition in DeCo~\cite{ma2025deco} and perceptual supervision in PixelGen~\cite{ma2026pixelgen}. We adopt pixel-space diffusion rather than VAE-latent diffusion because it avoids autoencoder bottlenecks and learned latent-space biases, providing a cleaner setting for studying co-denoising and representation alignment.

 \noindent\textbf{Representation alignment for diffusion training.}
A growing line of work studies how pretrained visual representations can improve diffusion training. Recent analyses~\cite{wang2025repa, singh2025irepa} show that diffusion models learn meaningful internal features, but these are often weaker or less structured than those of strong self-supervised vision encoders. REPA~\cite{wang2025repa} aligns intermediate diffusion features with pretrained representations such as DINOv2~\cite{oquab2023dinov2}, improving convergence and sample quality. Follow-up work studies which teacher properties matter most: iREPA~\cite{singh2025irepa} highlights spatial structure, while REPA-E~\cite{leng2025repa} extends REPA-style supervision to end-to-end latent diffusion training with the VAE. Recent results also suggest that REPA-style alignment is most beneficial early in training and may over-constrain the representation space if applied too rigidly~\cite{wang2025repa}. Motivated by this, we study representation alignment through co-denoising, compare auxiliary objectives beyond REPA, and introduce a stronger hybrid alternative.

\noindent\textbf{Visual co-denoising and joint generation across modalities.}
Recent work has increasingly explored \emph{joint denoising} or \emph{joint generation} of multiple signals to improve information transfer, controllability, and structural consistency. In image generation, Latent Forcing~\cite{baade2026latent} and ReDi~\cite{kouzelisboosting} jointly model image latents and semantic features. In video generation, VideoJAM~\cite{chefervideojam}, UDPDiff~\cite{yang2025unified}, and UnityVideo~\cite{huang2025unityvideo} jointly generate video with structured signals such as segmentation, depth, or flow. Similar ideas extend to audio--visual generation~\cite{wu2025does}, robotics and world modeling~\cite{zhong2025flowvla, bi2025motus, yang2025echomotion, dang2025syncmv4d, dangsvimo}, and multimodal sequence modeling~\cite{han2025tv2tv}. In contrast to these task-specific end-to-end designs, we provide a controlled study of visual co-denoising itself, isolating the architectural, guidance, loss, and calibration choices that make it effective and distilling them into a practical recipe for visual representation alignment.

%% file: sec/3_method.tex
\section{A Closer Look at Visual Co-Denoising}
\label{sec:method}

In this section, we first formalize visual co-denoising in \Cref{subsec:experiment_setup}, then conduct a systematic study of the key design choices that govern its effectiveness, including model architecture (\Cref{subsec:model_architecture}), unconditional prediction for CFG (\Cref{subsec:model_architecture_unconditional_generation}), auxiliary training objectives (\Cref{subsec:auxiliary_loss}), and feature calibration via rescaling (\Cref{subsec:rms_scaling_time_shift}). Starting from a standard pixel-space diffusion baseline (\eg{}, JiT~\cite{li2025back}), we use controlled ablations to isolate each component’s contribution and derive a practical recipe, introducing \emph{new designs tailored for visual co-denoising} along the way. Experiment setup details and additional ablations are deferred to Appendix Sec. A and Appendix Sec. B respectively.

\subsection{Co-Denoising Formulation}
\label{subsec:experiment_setup}
We formalize visual co-denoising within a unified framework.
Unlike standard pixel-space diffusion, which denoises only the image stream, co-denoising introduces an additional semantic feature stream from a pretrained visual encoder (\eg{}, DINOv2~\cite{oquab2023dinov2}). The core idea is to jointly denoise the pixel and semantic streams under a shared diffusion process, allowing the semantic stream to provide complementary supervision for semantically richer generation.

All experiments in this section follow the JiT~\cite{li2025back} ablation protocol on ImageNet 256$\times$256~\cite{deng2009imagenet}, training JiT-B/16 for 200 epochs. Unless otherwise specified, we adopt the JiT training configuration without hyperparameter tuning.

Concretely, we extend the $x$-prediction and $v$-loss formulation of JiT to jointly denoise pixels and pretrained semantic features. Let $\bm{x}$ denote the clean image and $\bm{d}$ denote its encoded patch-level semantic features. We sample independent Gaussian noise $\bm{\epsilon}_x,\bm{\epsilon}_d\sim\mathcal{N}(\bm{0},\bm{I})$ for the two streams. At diffusion time $t\in[0,1]$, the corresponding noised inputs are
\begin{align}
\label{eq:data_noise_interpolation}
\bm{z}_t^x &= t\,\bm{x} + (1-t)\,\bm{\epsilon}_x, &
\bm{z}_t^d &= t\,\bm{d} + (1-t)\,\bm{\epsilon}_d.
\end{align}
Given $(\bm{z}_t^x,\bm{z}_t^d,t,c)$, where $c$ denotes the class condition, the co-denoising model jointly predicts the clean targets for the pixel and semantic streams:
\begin{equation}
(\hat{\bm{x}}, \hat{\bm{d}})
=
f_{\bm{\theta}}(\bm{z}_t^x,\bm{z}_t^d,t,c),
\end{equation}
where $f_{\bm{\theta}}$ denotes the co-denoising model, which could be implemented as either a shared-backbone or dual-stream architecture depending on the design variant. Following JiT, we convert these clean predictions into velocity predictions,
\begin{align}
\hat{\bm{v}}_x &= (\hat{\bm{x}}-\bm{z}_t^x)/(1-t), &
\hat{\bm{v}}_d &= (\hat{\bm{d}}-\bm{z}_t^d)/(1-t),
\end{align}
and supervise them using the corresponding ground-truth velocities,
\begin{align}
\bm{v}_x &= \bm{x}-\bm{\epsilon}_x = (\bm{x}-\bm{z}_t^x)/(1-t), &
\bm{v}_d &= \bm{d}-\bm{\epsilon}_d = (\bm{d}-\bm{z}_t^d)/(1-t).
\end{align}
The final objective is a weighted sum of the pixels and semantic features $v$-losses:
\begin{equation}
\label{eq:loss}
\mathcal{L}_{\text{v-co}}
=
\mathbb{E}\Big[
\|\hat{\bm{v}}_x-\bm{v}_x\|_2^2
+
\lambda_d\,\|\hat{\bm{v}}_d-\bm{v}_d\|_2^2
\Big],
\end{equation}
where $\lambda_d$ controls the weight of the semantic stream. This formulation provides a unified testbed for studying the effects of architecture, guidance, auxiliary objectives, and feature calibration on representation alignment in co-denoising.

\begin{figure*}[t]
    \centering
    \includegraphics[width=0.99\linewidth]{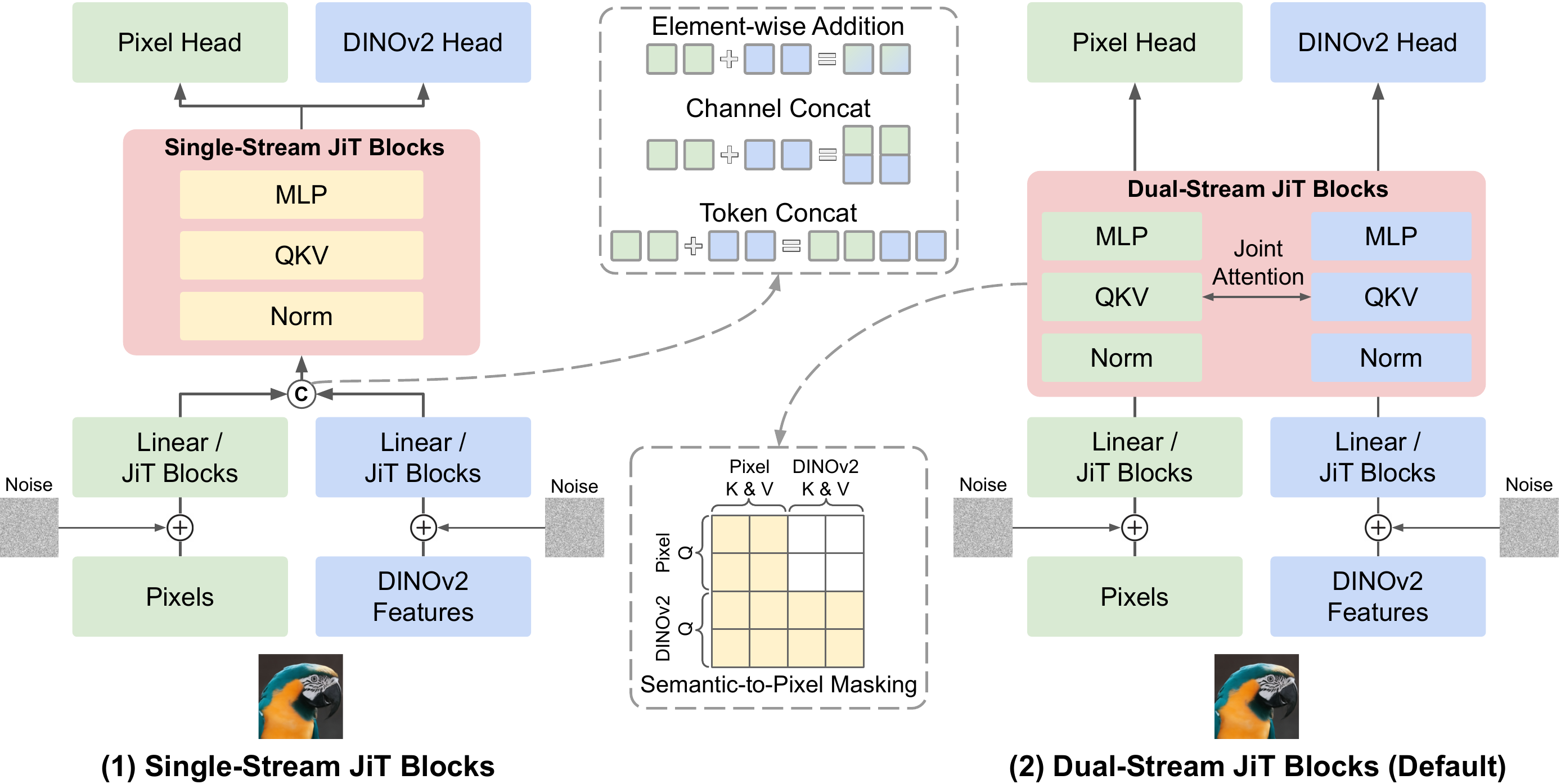}
    \caption{\textbf{Single-stream and dual-stream architectures for visual co-denoising.}
In the \emph{single-stream design} (left), noised pixels and DINOv2 features are fused after lightweight stream-specific preprocessing and then processed by shared JiT blocks. We study direct addition, channel concatenation, and token concatenation (see~\cref{subsec:model_architecture}). In the \emph{dual-stream design} (right), the two streams use separate normalization, MLP, and attention projections, while interacting through joint self-attention. A semantic-to-pixel attention mask is used to define the unconditional prediction for CFG (see~\cref{subsec:model_architecture_unconditional_generation}). Both designs use separate output heads for pixel and DINOv2 prediction.
}   
\label{fig:model_architecture_ablation}
\end{figure*}

\begin{table}[t]
\begin{center}
\setlength{\tabcolsep}{0.65em}
\renewcommand{\arraystretch}{1.0}
\caption{\textbf{Comparison of architectural designs for visual co-denoising.} We compare baseline backbones, single-stream fusion strategies, and dual-stream fusion variants with different allocations of feature-specific and shared/dual-stream blocks. All variants keep the pixel stream depth fixed at 12 JiT blocks for fair comparison. JiT-B/16$^{\ddagger}$ and JiT-B/16$^{\dagger}$ denote widened variants with hidden dimensions increased from 768 to 1024 and 1088, respectively, to match the parameter counts of the dual-stream models. Blue rows mark the stronger variants used in subsequent analysis. Following previous works~\cite{li2025back, baade2026latent}, we mainly use FID as reference.
}
\label{table:model_arch_ablation}
\scalebox{0.72}{
\begin{tabular}{clcccccc}
\hlineB{3}
 \multirow{2}{*}{} & \multirow{2}{*}{\textbf{Model}} & \multirow{2}{*}{\textbf{Backbone}} & \multirow{2}{*}{\textbf{\#Params}} & \textbf{\#Feature-Specific}  & \textbf{\#Shared/Dual-} & \multicolumn{2}{c}{\textbf{CFG=1.0}} \\
 \cline{6-7}
 & & & & \textbf{Blocks} & \textbf{Stream Blocks} & \textbf{FID$\downarrow$} & \textbf{IS$\uparrow$} \\ 
\hline
 \multicolumn{3}{l}{\textcolor{gray}{Baselines}} \\
(a) & JiT-B/16~\cite{li2025back}      & JiT-B/16            & 133M & - & -  & 32.54 & \textcolor{gray}{49.5}  \\
(b) &  JiT-B/16~\cite{li2025back}      & JiT-B/16$^{\dagger}$   & 261M & - & -  & 22.67 & \textcolor{gray}{69.9}  \\
(c) &  LatentForcing~\cite{baade2026latent} & JiT-B/16            & 156M & 2 & 10 & 13.06 & \textcolor{gray}{102.2}  \\
\hline
 \multicolumn{3}{l}{\textcolor{gray}{Single-Stream JiT Architecture}} \\
 (d) & DirectAddition& JiT-B/16            & 156M & 2 & 10 & 15.15 & \textcolor{gray}{103.4}  \\
 (e) & ChannelConcat & JiT-B/16            & 157M & 2 & 10 & 14.33 & \textcolor{gray}{107.7}  \\
          (f)  & TokenConcat & JiT-B/16            & 156M & 2 & 10 & 14.70 & \textcolor{gray}{103.8}  \\
         (g)  &  TokenConcat & JiT-B/16            & 177M & 4 & 8  & 12.59 & \textcolor{gray}{112.8}  \\
         (h)  & TokenConcat  & JiT-B/16            & 198M & 6 & 6  & {12.35} & \textcolor{gray}{116.7}  \\
         (i)  & TokenConcat  & JiT-B/16$^{\ddagger}$            & 265M & 6 & 6  & 9.74 &  \textcolor{gray}{129.45}  \\
\hline
\multicolumn{3}{l}{\textcolor{gray}{Dual-Stream JiT Architecture}}  \\
\rowcolor{lightblue}
(j) & TokenConcat & JiT-B/16            & 260M & 6 & 6  & 11.78 & {\textcolor{gray}{115.4}} \\
\rowcolor{lightblue}
(k) &  TokenConcat & JiT-B/16            & 260M & 4 & 8  & 11.40 & \textcolor{gray}{118.3} \\
\rowcolor{lightblue}
(l) & TokenConcat & JiT-B/16            & 260M & 2 & 10 & 10.24 & \textcolor{gray}{124.5} \\
\rowcolor{lightblue}
(m) & TokenConcat & JiT-B/16            & 260M & 0 & 12 & \textbf{8.86} & \textcolor{gray}{\textbf{132.8}}  \\
\hlineB{3}
\end{tabular}} 
\end{center}
\end{table}

\subsection{What Architecture Best Supports Visual Co-Denoising?}
\label{subsec:model_architecture}

We begin by studying how semantic features should be integrated into a pixel-space diffusion backbone for co-denoising. Our goal is to identify the \emph{architectural design that most effectively transfers information from pretrained semantic visual encoders to pixel features without limiting the expressiveness of the diffusion model}. To this end, we compare lightweight fusion within a largely shared backbone against more expressive designs that preserve feature-specific processing while enabling controlled cross-stream interaction. \cref{fig:model_architecture_ablation} illustrates the architectural variants, and \Cref{table:model_arch_ablation} summarizes the corresponding results.

\noindent\textbf{Baselines.}
We first report results for the original JiT-B/16 backbone~\cite{li2025back} and a widened variant that increases the hidden dimension from 768 to 1088, so as to match the parameter count of the dual-stream models introduced later. We also include Latent Forcing~\cite{baade2026latent} as a representative co-denoising baseline. For fair comparison, we keep the number of JiT blocks traversed by the pixel stream fixed across all variants, and maintain this setting throughout this subsection.

\noindent\textbf{Single-stream variants.}
We consider a shared-backbone setting (\cref{fig:model_architecture_ablation}, left) where pixel tokens $\bm{x}$ and semantic tokens $\bm{d}$ share most parameters. Within this setting, we compare three fusion strategies with Latent Forcing~\cite{baade2026latent} architecture:
\begin{itemize}[leftmargin=1.2em]
    \item \textbf{Direct Addition} (row (d)): Pixel tokens $\bm{x}\in\mathbb{R}^{n\times d_1}$ and semantic features $\bm{d}\in\mathbb{R}^{n\times d_2}$ are first projected into a shared hidden space $\mathbb{R}^{n\times d}$ via lightweight linear layers, then fused by \emph{element-wise addition} and passed through shared JiT blocks. The pixel and semantic streams have two separate output heads.
    \item \textbf{Channel-concatenation fusion} (row (e)): Pixel tokens $\bm{x}$ and semantic features $\bm{d}$ are concatenated along the channel dimension $\mathbb{R}^{n\times(d_1+d_2)}$, and then linearly projected to the hidden dimension $\mathbb{R}^{n\times d}$ of JiT blocks.
    \item \textbf{Token-concatenation fusion} (rows (f-i)): Instead of concatenating along the channel dimension, we concatenate $\bm{x}$ and $\bm{d}$ tokens along the sequence dimension $\mathbb{R}^{2n\times d}$ and input the combined token sequence into the JiT blocks.
\end{itemize}

\noindent\textbf{Dual-stream variants.}
Motivated by the limitations of heavily shared backbones, we further introduce a \emph{dual-stream} JiT architecture, illustrated on the right of \cref{fig:model_architecture_ablation}, in which the pixel and semantic streams maintain separate normalization layers, MLPs, and attention projections (\ie{}, Q/K/V), while interacting through joint self-attention. This design allows the model to adaptively determine \emph{where} and \emph{how} the two streams interact, while preserving dedicated processing pathways for each stream. 

\noindent\textbf{Analysis.}
As shown in \Cref{table:model_arch_ablation}, token-concatenation fusion outperforms direct addition and channel concatenation among the single-stream variants (rows (d)--(f)), suggesting that preserving feature-specific representations before interaction is preferable to early fusion in a shared space. 
Moreover, within token-concatenation, allocating more blocks to feature-specific processing consistently improves performance (rows (f)--(h)), indicating that excessive parameter sharing limits the model's ability to preserve semantic information. 
Finally, among the dual-stream variants, the fully dual-stream architecture (row (m)) achieves the best FID of 8.86 under a comparable number of trainable parameters (row (i) and rows (j)--(l)), showing that allowing the model to \emph{dynamically learn} cross-stream interaction at each block is more effective than imposing a fixed interaction pattern through a largely shared backbone. Therefore, we adopt the fully dual-stream architecture as the default model design in the remaining analysis.

\begin{questionbox}
\textbf{Finding 1:} Effective visual feature alignment requires preserving feature-specific processing while enabling flexible cross-stream interaction; token concatenation and a fully dual-stream JiT emerge as the preferred designs.
\end{questionbox}

\begin{table}[t]
\begin{center}
\setlength{\tabcolsep}{0.65em}
\renewcommand{\arraystretch}{1.0}
\caption{\textbf{Comparison of unconditional prediction designs for classifier-free guidance under co-denoising}. We report unguided results at CFG$=1.0$ and guided results at CFG$=2.9$, which is the default guided evaluation setting in JiT~\cite{li2025back}. 
}
\label{table:unconditional_generation}
\scalebox{0.71}{
\begin{tabular}{lllcccc}
\hlineB{3}
\multirow{2}{*}{\textbf{Uncond. Type}} & \multirow{2}{*}{\textbf{Cond. Pred.}} & \multirow{2}{*}{\textbf{Uncond. Pred.}} & \multicolumn{2}{c}{\textbf{CFG=1.0}} & \multicolumn{2}{c}{\textbf{CFG=2.9}}  \\  \cline{4-5} \cline{6-7}  
 & & & \textbf{FID$\downarrow$} & \textbf{IS$\uparrow$} & \textbf{FID$\downarrow$} & \textbf{IS$\uparrow$} \\  
\hline
 \multicolumn{3}{l}{\textcolor{gray}{Independently Drop Labels \& Semantic Features}} \\
 (a) Zero Embedding & $f_{\bm{\theta}}([\bm{x}_t, \bm{d}_t], y, t)$  & $f_{\bm{\theta}}([\bm{x}_t, \bm{0}], \emptyset, t)$  & 9.17 & \textcolor{gray}{126.3} & 6.69 & \textcolor{gray}{165.6} \\
 (b) Learnable \texttt{[null]} Token & $f_{\bm{\theta}}([\bm{x}_t, \bm{d}_t], y, t)$  & $f_{\bm{\theta}}([\bm{x}_t, \texttt{[null]}], \emptyset, t)$  & 9.37 & \textcolor{gray}{126.7} & 6.64 & \textcolor{gray}{165.7} \\
 (c) Bidirectional Cross-Stream Mask & $f_{\bm{\theta}}([\bm{x}_t, \bm{d}_t], y, t)$  & $f_{\bm{\theta}}([\bm{x}_t, \bm{d}_t], \emptyset, t)$  & 11.08 & \textcolor{gray}{101.9} & 7.17 & \textcolor{gray}{143.5} \\
  \rowcolor{lightblue}
 (d) Semantic-to-Pixel Mask & $f_{\bm{\theta}}([\bm{x}_t, \bm{d}_t], y, t)$  & $f_{\bm{\theta}}([\bm{x}_t, \bm{d}_t], \emptyset, t)$  & \textbf{7.28} & \textcolor{gray}{\textbf{136.8}} & \textbf{3.59} & \textcolor{gray}{\textbf{189.6}} \\
\hline
 \multicolumn{2}{l}{\textcolor{gray}{Jointly Drop Labels \& Semantic Features}} \\
 (e) Zero Embedding & $f_{\bm{\theta}}([\bm{x}_t, \bm{d}_t], y, t)$  & $f_{\bm{\theta}}([\bm{x}_t, \bm{0}], \emptyset, t)$  & 15.58 & \textcolor{gray}{98.8} & 24.75 & \textcolor{gray}{82.4} \\
 (f) Learnable \texttt{[null]} Token & $f_{\bm{\theta}}([\bm{x}_t, \bm{d}_t], y, t)$  & $f_{\bm{\theta}}([\bm{x}_t, \texttt{[null]}], \emptyset, t)$  & 10.80 & \textcolor{gray}{118.3} & 25.2 & \textcolor{gray}{88.3} \\
 (g) Bidirectional Cross-Stream Mask & $f_{\bm{\theta}}([\bm{x}_t, \bm{d}_t], y, t)$  & $f_{\bm{\theta}}([\bm{x}_t, \bm{d}_t], \emptyset, t)$  & 7.53 & \textcolor{gray}{129.1} & 5.66 & \textcolor{gray}{173.6} \\
  \rowcolor{lightblue}
   \rowcolor{lightblue}
 (h) Semantic-to-Pixel Mask & $f_{\bm{\theta}}([\bm{x}_t, \bm{d}_t], y, t)$  & $f_{\bm{\theta}}([\bm{x}_t, \bm{d}_t], \emptyset, t)$  & \textbf{5.62} & \textcolor{gray}{\textbf{158.5}} & \textbf{3.18} & \textcolor{gray}{\textbf{219.4}} \\
\hlineB{3}
\end{tabular}}
\end{center}
\end{table}

\subsection{How to Define Unconditional Prediction for CFG?}
\label{subsec:model_architecture_unconditional_generation}

To enable classifier-free guidance (CFG), the model must define an unconditional prediction, \ie{}, a prediction in which the conditioning signals are removed. In our co-denoising setting, this is nontrivial because the model is conditioned on both class labels and semantic features. Guided sampling combines the conditional and unconditional predictions in the pixel and semantic streams as
\begin{align}
\hat{\bm{v}}_x &= \hat{\bm{v}}_x^{\mathrm{uncond}} + s\left(\hat{\bm{v}}_x^{\mathrm{cond}}-\hat{\bm{v}}_x^{\mathrm{uncond}}\right),\quad
\hat{\bm{v}}_d &= \hat{\bm{v}}_d^{\mathrm{uncond}} + s\left(\hat{\bm{v}}_d^{\mathrm{cond}}-\hat{\bm{v}}_d^{\mathrm{uncond}}\right)
\end{align}
where $s$ denotes the CFG scale. Since guided generation depends critically on the quality of the unconditional branch, we next investigate \emph{how to define an effective unconditional prediction for CFG in the co-denoising setting}.

\noindent\textbf{Input-dropout baselines.}
Following prior work~\cite{baade2026latent, kouzelisboosting}, we first consider baseline unconditional predictions that drop conditioning inputs (semantic features and class labels) during training. Specifically, for semantic feature dropping, we use either \textbf{(1)} zeros or \textbf{(2)} a learnable \texttt{[null]} token to replace the semantic features. For each choice, we compare independent dropout of the class label and semantic features (rows (a)--(b)) against joint dropout (rows (e)--(f)).

\begin{wrapfigure}[10]{r}{0.45\linewidth}
  \centering
  \includegraphics[width=\linewidth]{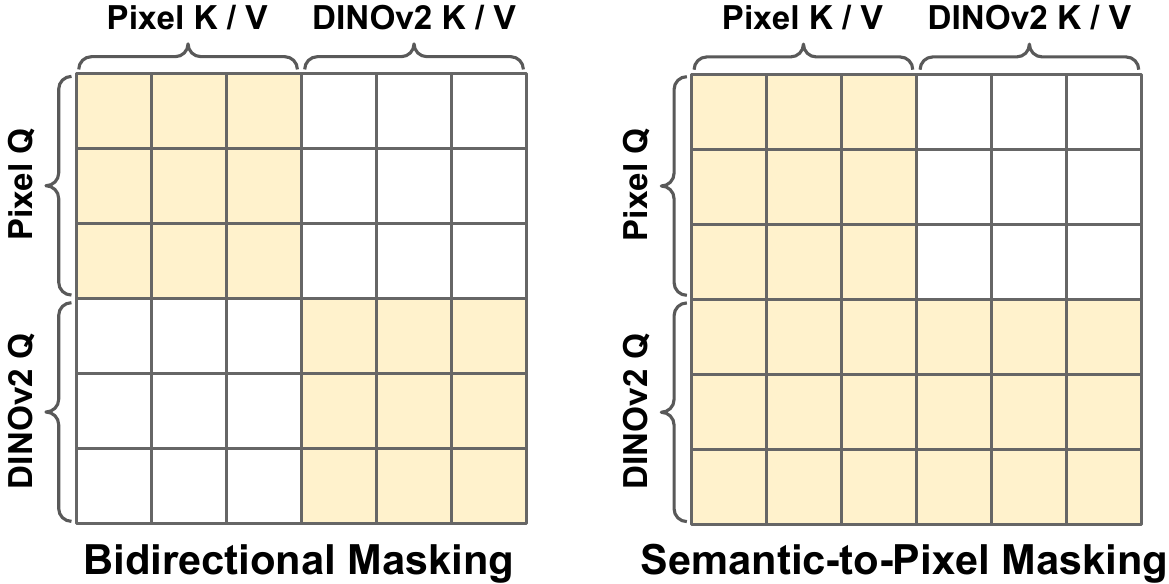}
  \caption{{Comparison of two attention-masking strategies.} 
  }
  \label{fig:attention_mask}
\end{wrapfigure}
\noindent\textbf{Attention mask between pixel and semantic features.}
Beyond input-level dropout, we leverage the dual-stream architecture to define a \emph{structurally unconditional} pathway. For unconditional samples, we apply \emph{semantic-to-pixel masking} (see \cref{fig:attention_mask}), which blocks cross-stream attention from the semantic stream to the pixel stream so that the pixel branch receives no semantic conditioning signal (rows (d) and (h)). We also study a symmetric variant, \emph{bidirectional cross-stream masking}, which blocks attention in both directions (rows (c) and (g)). These variants test whether unconditional prediction is better defined through explicit control of information flow rather than input-level corruption.

\noindent\textbf{Analysis.}
\Cref{table:unconditional_generation} first shows that under the baseline input-dropout strategy, independently dropping the class label and semantic features (rows (a)--(b)) performs substantially better than jointly dropping them (rows (e)--(f)). We hypothesize that jointly dropping both conditions makes the pixel-space guidance direction,
$\Delta_x=\hat{\bm{v}}_x^{\mathrm{cond}}-\hat{\bm{v}}_x^{\mathrm{uncond}}$,
a poorly calibrated estimate of the desired conditional guidance signal, which is then amplified by CFG scaling. In contrast, independent dropout exposes the model to partially conditioned cases and thus appears to improve the robustness of the learned guidance direction.

More importantly, explicitly defining the unconditional pathway through \emph{structural} masking (rows (c)--(d)) is markedly more effective than input-level dropout (rows (a)--(b)) under independent dropout, suggesting that blocking semantic information from reaching the pixel branch yields a more reliable unconditional prediction. Among the structural variants, masking only the semantic-to-pixel pathway (row (d)) performs best, indicating that unconditional generation only requires removing semantic influence on the pixel output, while preserving the reverse pixel-to-semantic interaction remains beneficial. For structural masking, jointly dropping labels and semantic features (rows (g)--(h)) outperforms independent dropout (rows (c)--(d)), suggesting that once the unconditional branch is defined structurally, removing all conditioning sources during training better matches inference-time behavior. 

\begin{questionbox}
\textbf{Finding 2:} For CFG under co-denoising, the most effective unconditional design is structural semantic-to-pixel masking with joint dropout of conditioning signals during training.
\end{questionbox}

\subsection{Which Auxiliary Loss Best Improves Co-Denoising?}
\label{subsec:auxiliary_loss}

The default V-Co objective in \cref{eq:loss} supervises both streams through the co-denoising $v$-loss, but it mainly enforces local target matching and may not fully capture higher-level semantic alignment. We therefore study \emph{which auxiliary objectives provide the most effective complementary supervision in representation space}, and ultimately design a hybrid loss that better combines their strengths.

\begin{wrapfigure}[12]{r}{0.38\linewidth}
  \centering
  \includegraphics[width=\linewidth]{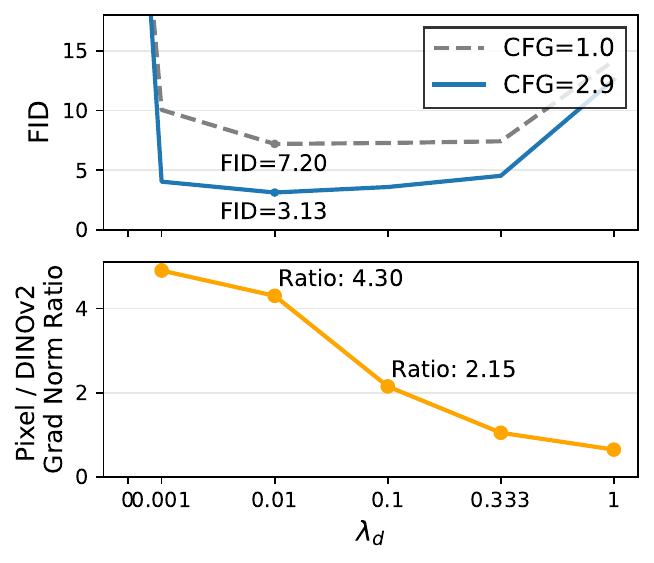}
  \caption{
  Influence of $\lambda_d$.
  }
  \label{fig:feature_vloss}
\end{wrapfigure}
\noindent\textbf{Balancing pixel and semantic losses.}
Before exploring additional auxiliary objectives, we first tune the relative weight of the semantic-stream loss through $\lambda_d$ in \cref{eq:loss}. As shown in~\cref{fig:feature_vloss}, $\lambda_d\in\{0.01, 0.1\}$ gives the best FID. Under these settings, the average parameter-gradient norm in the pixel branch is approximately $4\times$ and $2\times$ that of the semantic branch, respectively. This suggests that semantic supervision is most effective when it provides meaningful guidance while remaining secondary to the primary pixel-space objective.

\noindent\textbf{REPA loss~\cite{leng2025repa}.}
On top of the co-denoising objective, we further consider a REPA-style representation alignment loss on the pixel branch. Concretely, let $\bm{h}_{\ell}^x$ denote the intermediate hidden representation of the pixel branch at layer $\ell$. We align this hidden state to the representation of the ground-truth image $\bm{x}$ extracted by a frozen pretrained DINOv2 visual encoder $\phi(\cdot)$. 
The auxiliary objective is defined as
$\mathcal{L}_{\mathrm{REPA}}=\big\|g(\bm{h}_{\ell}^x)-\phi(\bm{x})\big\|_2^2$,
where $g(\cdot)$ denotes a lightweight MLP projector used to map the intermediate hidden state to the encoder feature space.

\noindent\textbf{Perceptual loss in semantic feature space.}
We also consider a perceptual loss~\cite{johnson2016perceptual, ma2026pixelgen} in the pretrained semantic feature space. Specifically, given the predicted clean image $\hat{\bm{x}}$ and the ground-truth image $\bm{x}$, we extract their features using a frozen pretrained DINOv2 visual encoder $\phi(\cdot)$ and minimize their discrepancy:
$\mathcal{L}_{\mathrm{perc}}=\|\phi(\hat{\bm{x}})-\phi(\bm{x})\|_2^2$.
Unlike REPA, which aligns intermediate hidden states, this loss directly supervises the predicted image in semantic feature space.

\noindent\textbf{Drifting loss in semantic feature space.}
Drifting loss~\cite{deng2026generative} was recently proposed for single-step image generation to move the generated distribution toward the real one. Unlike diffusion $v$-prediction, REPA, and perceptual losses, which impose pairwise supervision between generated and real samples, drifting loss operates at the distribution level. To study its effectiveness in multi-step diffusion, we implement it in DINO feature space. Let $\phi(\cdot)$ denote a frozen DINOv2 encoder, let $\hat{\bm{x}}$ be the predicted clean image from the pixel branch, and define $\bm{u}=\phi(\hat{\bm{x}})$. We construct a drifting field as follows:
\begin{align}
V(\bm{u}) &= V^{+}(\bm{u}) - V^{-}(\bm{u}), \\
V^{+}(\bm{u}) &= \frac{1}{Z_{+}(\bm{u})}\,\mathbb{E}_{\bm{x}^{+}\sim p_{\mathrm{data}}}
\big[k(\bm{u},\phi(\bm{x}^{+}))(\phi(\bm{x}^{+})-\bm{u})\big], \\
V^{-}(\bm{u}) &= \frac{1}{Z_{-}(\bm{u})}\,\mathbb{E}_{\bm{x}^{-}\sim p_{\mathrm{gen}}}
\big[k(\bm{u},\phi(\bm{x}^{-}))(\phi(\bm{x}^{-})-\bm{u})\big],
\label{eq:v_minus}
\end{align}
where $p_{\mathrm{data}}$ and $p_{\mathrm{gen}}$ denote the real and generated image distributions respectively. $k(\bm{a},\bm{b})=\exp(-\|\bm{a}-\bm{b}\|_2^2/\tau)$ is a similarity kernel and $Z_{+},Z_{-}$ normalize the kernel weights. The drifting loss is defined as
$\mathcal{L}_{\mathrm{drift}}=\|\bm{u}-\mathrm{sg}(\bm{u}+V(\bm{u}))\|_2^2$, where $\mathrm{sg}(\cdot)$ denotes stop-gradient.

\begin{table}[t]
\begin{center}
\setlength{\tabcolsep}{0.65em}
\renewcommand{\arraystretch}{1.0}
\caption{\textbf{Comparison of auxiliary losses applied to V-Co.} We report unguided results at CFG$=1.0$ and guided results at the best CFG selected from a sweep over $[1.5,\,5.5]$ for each method. All results here are obtained after 300 epochs of training.
}
\label{table:loss}
\scalebox{0.78}{
\begin{tabular}{lllcc}
\hlineB{3}
\multirow{2}{*}{\textbf{Aux. Loss Type}} & \multicolumn{2}{c}{\textbf{Unguided (CFG=1.0)}} & \multicolumn{2}{c}{\textbf{Guided (Best CFG$>1.0$)}}  \\  \cline{2-3} \cline{4-5}  
 & \textbf{FID$\downarrow$} & \textbf{IS$\uparrow$} & \textbf{FID$\downarrow$} & \textbf{IS$\uparrow$} \\  
\hline
 Baseline: V-Co w/ default V-Pred Loss & 5.38 & \textcolor{gray}{153.6} & 2.96 & \textcolor{gray}{206.6} \\
 (a) V-Co + REPA Loss & 5.63 & \textcolor{gray}{149.4} & 2.91 {\color{blue}$(0.05\downarrow)$} & \textcolor{gray}{202.8} \\
 (b) V-Co + Perceptual Loss & \textbf{4.28} & \textcolor{gray}{177.6} & 2.73 {\color{blue}$(0.23\downarrow)$} & \textcolor{gray}{228.5} \\
 (c) V-Co + Drifting Loss  & 4.86 & \textcolor{gray}{164.3} & 2.85 {\color{blue}$(0.11\downarrow)$} & \textcolor{gray}{211.5} \\
  \rowcolor{lightblue}
 (d) V-Co + Perceptual-Drifting Hybrid Loss & {4.44} & \textcolor{gray}{\textbf{189.0}} & \textbf{2.44} {\color{blue}$(0.52\downarrow)$} & \textcolor{gray}{\textbf{249.9}} \\
\hlineB{3}
\end{tabular}}
\end{center}
\end{table}

\noindent\textbf{Perceptual-Drifting Hybrid loss.}
Perceptual loss and drifting loss provide two complementary forms of supervision. Perceptual loss encourages \emph{instance-level} semantic fidelity by pulling each generated image toward the semantic feature of its paired ground-truth target, while drifting loss promotes \emph{distributional coverage} by discouraging generated features from collapsing toward dense regions of the generated distribution. Motivated by this complementarity, we propose a hybrid objective that formulates perceptual alignment as a positive vector field and drifting-based repulsion as a negative correction.

Unlike the original drifting loss, we replace its positive real-distribution term with the current sample’s \emph{paired perceptual field} while retaining the generated-distribution term as a negative correction. This is better suited to multi-step denoising, where each noisy input is paired with a specific ground-truth image.

Specifically, we define the positive perceptual field as:
\begin{equation}
V_{\mathrm{pos}}(\bm{u}_i) = \phi(\bm{x}_i) - \phi(\hat{\bm{x}}_i),
\end{equation}
which pulls the generated sample toward the semantic feature of its paired ground-truth target.
The negative field computes repulsion from nearby generated samples within the same class. For each sample $i$, we compute normalized kernel weights over other samples $j \neq i$ of the same class:
\begin{equation}
\alpha_{ij} = \frac{\exp\left(-\|\phi(\hat{\bm{x}}_i) - \phi(\hat{\bm{x}}_j)\|^2 / \tau_{\mathrm{rep}}\right)}{\sum_{k \neq i} \exp\left(-\|\phi(\hat{\bm{x}}_i) - \phi(\hat{\bm{x}}_k)\|^2 / \tau_{\mathrm{rep}}\right)},
\end{equation}
where $\tau_{\mathrm{rep}}$ is the repulsion temperature. Note that $\sum_{j \neq i} \alpha_{ij} = 1$.
The repulsion direction points from sample $i$ toward the weighted centroid of its neighbors:
\begin{equation}
V_{\mathrm{neg}}(\bm{u}_i) = \sum_{j \neq i} \alpha_{ij} \phi(\hat{\bm{x}}_j) - \phi(\hat{\bm{x}}_i).
\end{equation}
To adaptively balance attraction and repulsion, we introduce a \emph{similarity-based gating} mechanism based on how close the generated feature is to its target:
\begin{equation}
s_i = \exp\left(-\frac{\|\phi(\hat{\bm{x}}_i) - \phi(\bm{x}_i)\|^2}{\tau_{\mathrm{gate}}}\right),
\label{eq:similarity_based_gating}
\end{equation}
where $\tau_{\mathrm{gate}}$ is a temperature parameter controlling the sensitivity of the gate. We then combine the two fields into a hybrid field:
\begin{equation}
V_{\mathrm{hyb}}(\bm{u}_i) = s_i \cdot V_{\mathrm{pos}}(\bm{u}_i) - (1 - s_i) \cdot V_{\mathrm{neg}}(\bm{u}_i).
\label{eq:hybrid_field}
\end{equation}
Intuitively, when the generated feature is far from the target ($s_i \approx 0$), repulsion dominates to prevent mode collapse; when the generated feature is close to the target ($s_i \approx 1$), pure attraction ensures clean convergence.

The final objective is:
\begin{equation}
\mathcal{L}
=
\mathcal{L}_{\text{v-co}}
+
\lambda_{\mathrm{hyb}}\,\mathcal{L}_{\mathrm{hyb}}
=
\mathcal{L}_{\text{v-co}}
+
\lambda_{\mathrm{hyb}}
\left\|
\bm{u}_i
-
\mathrm{sg}\big(\bm{u}_i + V_{\mathrm{hyb}}(\bm{u}_i)\big)
\right\|_2^2.
\end{equation}

\begin{wrapfigure}[12]{r}{0.45\linewidth}
  \centering
  \includegraphics[width=\linewidth]{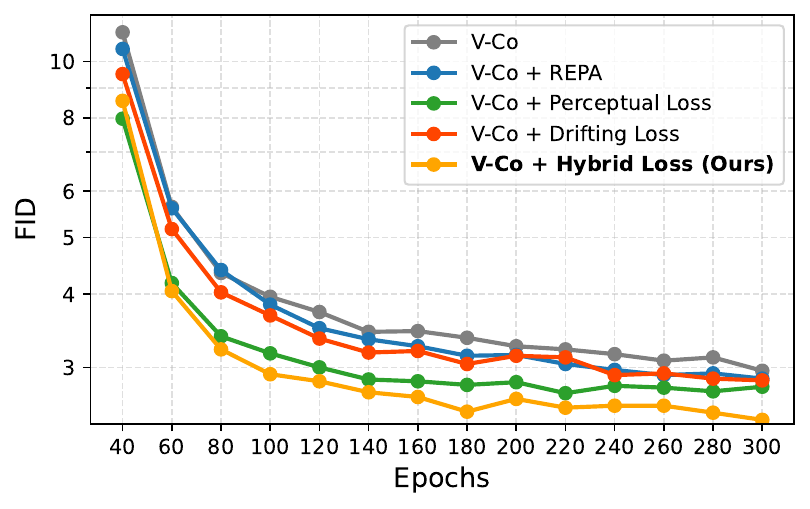}
  \caption{Comparison of guided FID.}
  \label{fig:aux_loss}
\end{wrapfigure}
\noindent\textbf{Analysis.}
\Cref{table:loss} and~\cref{fig:aux_loss} quantify the impact of different auxiliary objectives on co-denoising. REPA provides only a marginal guided improvement over the default V-Co objective (FID 2.91 vs.\ 2.96 for CFG$>1$), suggesting limited benefit from intermediate alignment and potential constraints on model's expressiveness~\cite{wang2025repa}. Perceptual loss yields a larger gain (FID 2.73), indicating effective instance-level semantic alignment, while drifting loss provides a smaller improvement (FID 2.85) but contributes distribution-level supervision. Combining these complementary signals, our hybrid objective achieves the best guided result (FID 2.44), suggesting that instance-level alignment is most effective when paired with explicit distribution-level correction.

\begin{questionbox}
\textbf{Finding 3:} For visual co-denoising, auxiliary losses work best when combining instance-level alignment with distribution-level regularization.
\end{questionbox}

\begin{table}[t]
\begin{center}
\setlength{\tabcolsep}{0.65em}
\renewcommand{\arraystretch}{1.0}
\caption{\textbf{Feature rescaling \vs{} noise-schedule shifting in V-Co.} We compare our default V-Co model with RMS scaling against variants \emph{without RMS scaling} and \emph{with noise-schedule shifting}. We report unguided results at CFG$=1.0$ and guided results at the best CFG selected from a sweep over $[1.5,\,5.5]$ for each method.}
\label{table:rms_scaling}
\scalebox{0.77}{
\begin{tabular}{lllcc}
\hlineB{3}
\multirow{2}{*}{\textbf{Model}} & \multicolumn{2}{c}{\textbf{Unguided (CFG=1.0)}} & \multicolumn{2}{c}{\textbf{Guided (Best CFG$>1.0$)}}  \\  \cline{2-3} \cline{4-5}  
 & \textbf{FID$\downarrow$} & \textbf{IS$\uparrow$} & \textbf{FID$\downarrow$} & \textbf{IS$\uparrow$} \\  
\hline
 V-Co w/o rms scaling  & 9.12 & \textcolor{gray}{188.6} & 5.28& \textcolor{gray}{150.4} \\
 V-Co w/o rms scaling + noise schedule shifting  & \textbf{4.81} & \textcolor{gray}{\textbf{205.6}} & 2.93 & \textcolor{gray}{\textbf{272.4}} \\
 \rowcolor{lightblue}
 V-Co w/ rms scaling (default)  & 5.38 & \textcolor{gray}{177.0} & \textbf{2.52} & \textcolor{gray}{242.6} \\
\hlineB{3}
\end{tabular}}
\end{center}
\end{table}

\subsection{How Should Semantic Features Be Calibrated for Co-Denoising?}
\label{subsec:rms_scaling_time_shift}

Before concluding the recipe, we consider \emph{how to calibrate the semantic stream relative to the pixel stream during co-denoising}. Since the two streams lie in different representation spaces and can have very different signal scales, using the same diffusion timestep may induce mismatched denoising difficulty, leading to imbalanced optimization. This motivates two natural strategies: rescaling the semantic features to match the pixel-space signal level, or equivalently shifting the semantic-stream diffusion schedule~\cite{baade2026latent}. As we show below, these two formulations are equivalent in signal-to-noise ratio (SNR).

\noindent\textbf{SNR matching via feature rescaling.}
As defined in~\cref{eq:data_noise_interpolation}, pixels and semantic features share the same timestep $t\in[0,1]$. Under this flow-matching parameterization, denoising difficulty is governed by:

\begin{equation}
    \mathrm{SNR}(t)\propto\frac{t^2\,\mathbb{E}\|\bm{s}\|^2}{(1-t)^2\,\mathbb{E}\|\bm{\epsilon}\|^2},
    \label{eq:snr_def}
\end{equation}
where $\bm{s}$ is the clean signal and $\bm{\epsilon}\sim\mathcal{N}(\bm{0},\bm{I})$ is the injected noise. Since $t$ is shared and the noise scale is fixed, matching denoising difficulty reduces to matching signal scale. We therefore rescale semantic features to match the pixel SNR:
\begin{equation}
    \bm{d}'=\alpha\bm{d},\qquad
    \alpha=\frac{\mathrm{RMS}_{\bm{x}}}{\mathrm{RMS}_{\bm{d}}}.
\end{equation}

\noindent\textbf{Equivalent SNR matching via timestep shifting.}
Equivalently, one can keep the semantic feature $\bm{d}$ fixed and instead shift its timestep from $t$ to $t'$ such that
$\mathrm{SNR}_{\bm{d}}(t')=\mathrm{SNR}_{\bm{d}'}(t)$.
Using \cref{eq:snr_def}, this gives
$\frac{t'}{1-t'}=\alpha\,\frac{t}{1-t}$,
and thus
\begin{equation}
    t'=\frac{\alpha t}{1+(\alpha-1)t}.
    \label{eq:time_shift}
\end{equation}
Therefore, rescaling the semantic features by $\alpha$ is SNR-equivalent to applying a shifted diffusion schedule to the unscaled semantic stream.

\noindent\textbf{Analysis.} In \Cref{table:rms_scaling}, we compare our default V-Co model against variants \emph{without RMS scaling} and \emph{with noise-schedule shifting} in place of RMS scaling. Removing RMS scaling substantially worsens performance relative to our default setting (guided FID: 5.28 \vs{} \ 2.52). Replacing RMS scaling with noise-schedule shifting gives broadly comparable overall results, but yields worse guided FID (2.93 \vs{} \ 2.52), consistent with the SNR-based equivalence discussed above. In practice, we adopt RMS scaling because of its simplicity and strong performance.

\begin{questionbox}
\textbf{Finding 4:} Effective co-denoising requires calibrating the semantic and pixel streams to comparable denoising difficulty. RMS-based feature rescaling provides a simple and practical solution.
\end{questionbox}

%% file: sec/4_experiments.tex
\section{Full Recipe and SoTA Comparison}
\label{sec:experiments}

\begin{table*}[t]
\begin{center}
\setlength{\tabcolsep}{0.55em}
\renewcommand{\arraystretch}{1.0}
\caption{\textbf{Reference results on ImageNet 256$\times$256.} FID and IS of 50K samples are evaluated. The ``pre-training'' columns list the external models required to obtain the results. The \#Params column reports the parameter count of the denoising model and the decoder (if used). Following previous works~\cite{li2025back, baade2026latent}, we mainly use FID as reference.}
\label{table:imagenet_256_comparison}
\scalebox{0.79}{
\begin{tabular}{lcc|cc|cc}
\hlineB{3}
\multirow{2}{*}{\textbf{ImageNet $256\times256$}} & \multicolumn{2}{c|}{\textbf{Pre-training}} & \multirow{2}{*}{\textbf{\#Params}}  & \multirow{2}{*}{\textbf{\#Epochs}}  & \multirow{2}{*}{\textbf{FID$\downarrow$}} & \multirow{2}{*}{\textbf{IS$\uparrow$}} \\
& \textbf{Tokenizer} & \textbf{SSL Encoder} & & &   \\
\hline

\multicolumn{2}{l}{\textcolor{gray}{Latent-space Diffusion}} \\
DiT-XL/2~\cite{peebles2023scalable} & SD-VAE & - & 675+49M & 1400  & 2.27 & {\textcolor{gray}{278.2}} \\
SiT-XL/2~\cite{ma2024sit} & SD-VAE & - & 675+49M & 1400 & 2.06 &  {\textcolor{gray}{277.5}} \\
REPA~\cite{wang2025repa}, SiT-XL/2 & SD-VAE & DINOv2 & 675+49M & 800  & 1.42 & {\textcolor{gray}{305.7}} \\
LightningDiT-XL/2~\cite{yao2025vavae} & VA-VAE & DINOv2 & 675+49M & 800  & 1.35 & {\textcolor{gray}{295.3}} \\
DDT-XL/2~\cite{wang2025ddt} & SD-VAE & DINOv2 & 675+49M & 400 &  1.26 & {\textcolor{gray}{310.6}} \\
RAE~\cite{zheng2025diffusion}, DiT$^{\mathrm{DH}}$-XL/2 & RAE & DINOv2 & 839+415M & 800 & \textbf{1.13} & {\textcolor{gray}{262.6}} \\

\hline
\multicolumn{2}{l}{\textcolor{gray}{Pixel-space (non-diffusion)}} \\
JetFormer~\cite{tschannenjetformer} & - & - & 2.8B & - & 6.64 & {\textcolor{gray}{-}} \\
FractalMAR-H~\cite{li2025fractal} & - & - & 848M & 600 & 6.15 & {\textcolor{gray}{\textbf{348.9}}} \\

\hline
\multicolumn{2}{l}{\textcolor{gray}{Pixel-space Diffusion}} \\
ADM-G~\cite{dhariwal2021diffusion} & - & - & 554M & 400 & 4.59 & {\textcolor{gray}{186.7}} \\
RIN~\cite{jabri2023scalable} & - & - & 410M &  - & 3.42 & {\textcolor{gray}{182.0}} \\
SiD~\cite{hoogeboom2023simple}, UViT/2 & - & - & 2B & 800 & 2.44 & {\textcolor{gray}{256.3}} \\
VDM++, UViT/2 & - & - & 2B & 800 & 2.12 & {\textcolor{gray}{267.7}} \\
SiD2~\cite{hoogeboom2025simpler}, UViT/2 & - & - & N/A & - & 1.73 & {\textcolor{gray}{-}} \\
SiD2~\cite{hoogeboom2025simpler}, UViT/1 & - & - & N/A & - & {1.38} & {\textcolor{gray}{-}} \\
PixelFlow~\cite{chen2025pixelflow}, XL/4 & - & - & 677M & 320 & 1.98 & {\textcolor{gray}{282.1}} \\
PixNerd~\cite{wang2025pixnerd}, XL/16 & - & DINOv2 & 700M & 160 & 2.15 & {\textcolor{gray}{297.0}} \\
DeCo-XL/16~\cite{ma2025deco} & - & DINOv2 & 682M &  600 & 1.69 & {\textcolor{gray}{304.0}} \\
PixelGen-XL/16~\cite{ma2026pixelgen} & - & DINOv2 & 676M &  160 & 1.83 & {\textcolor{gray}{293.6}} \\
ReDi~\cite{kouzelisboosting}, SiT-XL/2 & - & DINOv2 & 675M &   350 & 1.72 & {\textcolor{gray}{278.7}} \\
ReDi~\cite{kouzelisboosting}, SiT-XL/2 & - & DINOv2 & 675M &  800  & 1.61 & {\textcolor{gray}{295.1}} \\
Latent Forcing~\cite{baade2026latent}, JiT-B/16 & - & DINOv2 & 465M  & 200  & 2.48 & {\textcolor{gray}{-}} \\
JiT-B/16~\cite{li2025back} & - & - & 131M &  600 & 3.66 & {\textcolor{gray}{275.1}} \\
JiT-L/16~\cite{li2025back} & - & - & 459M &  600 & 2.36 & {\textcolor{gray}{298.5}} \\
JiT-H/16~\cite{li2025back} & - & - & 953M &  600 & 1.86 & {\textcolor{gray}{303.4}} \\
JiT-G/16~\cite{li2025back} & - & - & 2B & 600 & 1.82 & {\textcolor{gray}{292.6}} \\
\hline
 \rowcolor{lightblue}
V-Co-B/16 & - & DINOv2 & 260M &  {\textcolor{blue}{200}} & 2.52 & {\textcolor{gray}{242.6}} \\
 \rowcolor{lightblue}
V-Co-L/16 & - & DINOv2 & 918M &  {\textcolor{blue}{200}} & 2.10 & {\textcolor{gray}{243.0}} \\
 \rowcolor{lightblue}
V-Co-H/16 & - & DINOv2 & 1.9B &  {\textcolor{blue}{200}} & 1.85 & {\textcolor{gray}{246.5}}    \\
\hline
 \rowcolor{lightblue}
V-Co-B/16 & - & DINOv2 & 260M &  {\textcolor{red}{600}} & 2.35 & {\textcolor{gray}{261.1}} \\
 \rowcolor{lightblue}
V-Co-L/16 & - & DINOv2 & 918M &  {\textcolor{red}{500}} & 1.72 & {\textcolor{gray}{245.3}} \\
 \rowcolor{lightblue}
V-Co-H/16 & - & DINOv2 & 1.9B &  \textcolor{red}{300} & 1.71 & {\textcolor{gray}{263.3}} \\
\hlineB{3}
\end{tabular}}
\end{center}
\end{table*}

We combine the best-performing ablation choices into a practical visual co-denoising (\textbf{V-Co}) recipe: a fully dual-stream JiT backbone (\cref{subsec:model_architecture}), structural semantic-to-pixel masking with joint dropout for CFG (\cref{subsec:model_architecture_unconditional_generation}), a perceptual-drifting hybrid loss (\cref{subsec:auxiliary_loss}), and RMS-based feature rescaling (\cref{subsec:rms_scaling_time_shift}). These components are complementary: the architecture determines \emph{where} streams interact, masking defines \emph{how} guidance is formed, the hybrid loss specifies \emph{what} semantic supervision to apply, and RMS scaling matches denoising difficulty.

\begin{table*}[t]
\begin{center}
\setlength{\tabcolsep}{0.50em}
\renewcommand{\arraystretch}{1.0}
\caption{\textbf{Computational efficiency comparison with JiT.} 
We report trainable parameters, per-sample training GFLOPs, an epoch-normalized training-compute proxy, and inference throughput. For V-Co, total training GFLOPs include both the denoising Transformer and the frozen DINOv2 encoder used to extract semantic features during training. Inference throughput is measured without the DINOv2 encoder, since V-Co does not require the external encoder at sampling time.}
\label{table:efficiency_comparison}
\scalebox{0.78}{
\begin{tabular}{lcccccc}
\hlineB{3}
\multirow{2}{*}{\textbf{Model}} 
& \multirow{2}{*}{\textbf{\#Params}} 
& \multicolumn{3}{c}{\textbf{Train GFLOPs}}  
& \multirow{2}{*}{\textbf{Imgs/sec$\uparrow$}} 
& \multirow{2}{*}{\textbf{Epochs / FID$\downarrow$}} \\
\cline{3-5}
& 
& \textbf{DiT} 
& \textbf{DINO Enc.} 
& \textbf{Total$\downarrow$} 
& 
& \\
\hline
JiT-L/16~\cite{li2025back} 
& 459M 
& 89 
& - 
& 89 
& 0.329 
& 600 / 2.36 \\
\rowcolor{lightblue}
V-Co-B/16 
& 260M 
& 53 
& 35 
& 88 
& \textbf{0.484} 
& 600 / \textbf{2.35} \\
\hline
JiT-H/16~\cite{li2025back} 
& 953M 
& 183 
& - 
& 183 
& 0.175 
& 600 / 1.86 \\
JiT-G/16~\cite{li2025back} 
& 2.0B 
& 384 
& - 
& 384 
& 0.089 
& 600 / 1.82 \\
\rowcolor{lightblue}
V-Co-L/16 
& 918M 
& 183 
& 35 
& 218 
& \textbf{0.095} 
& 500 / \textbf{1.72} \\
\hlineB{3}
\end{tabular}}
\end{center}
\end{table*}

We next evaluate this full recipe against prior SoTA methods on ImageNet~\cite{deng2009imagenet} $256\times256$, including latent-space diffusion models, and pixel-space methods. As shown in~\Cref{table:imagenet_256_comparison}, V-Co achieves strong performance among pixel-space diffusion models. 
Notably, V-Co-B/16 with only 260M parameters, matches JiT-L/16 with 459M parameters (FID 2.35 \vs{} 2.36). V-Co-L/16 and V-Co-H/16, trained for 500 and 300 epochs respectively, outperform JiT-G/16 with 2B parameters (FID 1.71 \vs{} 1.82) and other strong pixel-diffusion methods. In addition, our method with simple RMS scaling matches or outperforms Latent Forcing~\cite{baade2026latent}, which relies on separate noise schedules for pixels and DINOv2 features.
These results demonstrate that a carefully designed co-denoising recipe is both effective and scalable for representation-aligned pixel-space generation.

\noindent\textbf{Computational efficiency.}
\TODO{We further compare training GFLOPs and inference throughput in Table~\ref{table:efficiency_comparison}. Although V-Co introduces an additional DINOv2 encoder during training, this cost is largely offset by the improved parameter efficiency of the co-denoising backbone. Specifically, V-Co-B achieves nearly identical total training GFLOPs to JiT-L (88 vs. 89), while matching its generation quality (FID 2.35 vs. 2.36) and providing higher inference throughput (0.484 vs. 0.329 images/s). V-Co-L improves over JiT-H/G in FID (1.72 vs. 1.86/1.82), while requiring substantially fewer total training GFLOPs than JiT-G (218 vs. 384) and slightly exceeding its inference throughput (0.095 vs. 0.089 images/s). 
These results suggest that visual co-denoising improves not only generation quality, but also the quality--compute trade-off of pixel-space diffusion models.}

%% file: sec/5_conclusion.tex
\section{Conclusion}
\label{sec:conclusion}

We presented V-Co, a visual co-denoising framework for representation-aligned pixel-space generation. We identify four key ingredients for effective co-denoising: a fully dual-stream backbone, structural semantic-to-pixel masking for classifier-free guidance, a perceptual-drifting hybrid loss, and RMS-based feature rescaling. Together, they form a simple and scalable recipe that improves semantic alignment and generative quality, with strong ImageNet results and clear scalability with model size and training duration. We hope this work can inspire future research on co-denoising and representation-aligned generative modeling.

\section*{Acknowledgment}
This work was supported by ONR Grant N00014-23-1-2356, ARO Award W911N-F2110220, DARPA ECOLE Program No. HR00112390060, NSF-AI Engage Institute DRL-2112635 and NVIDIA Academic Grant Program. The views contained in this article are
those of the authors and not of the funding agency.

%% file: sec/99_appendix.tex
\clearpage
\setcounter{page}{1}

\appendix

\section*{Appendix}

\section{Experiment Setup Details}
\label{subsec:appendix_experiment_setup_details}

We report the hyper-parameters of the final model used for evaluation in~\Cref{table:vco_configs}.
Throughout Sec. 3, these design choices are introduced progressively and tuned step by step toward the final configuration.

\begin{table}[h]
\begin{center}
\setlength{\tabcolsep}{0.65em}
\renewcommand{\arraystretch}{1.0}
\caption{\textbf{Configurations of experiments.} Architecture, feature pre-processing, training, and sampling settings for V-Co-B/L/H. We color the newly added hyper-parameters on top of JIT~\cite{li2025back} codebase in \colorbox{lightblue}{light blue}.}
\label{table:vco_configs}
\scalebox{0.85}{
\begin{tabular}{lccc}
\hlineB{3}
 & \textbf{V-Co-B} & \textbf{V-Co-L} & \textbf{V-Co-H} \\
\hline

\multicolumn{2}{l}{\textcolor{gray}{Architecture}} \\
Depth                  & 12   & 24   & 32    \\
Hidden dim             & 768  & 1024 & 1280  \\
Heads                  & 12   & 16   & 16    \\
Image size             & 256  & {256} & 256 \\
Patch size             & 16 & 16 & 16 \\
Bottleneck             & 128 & 128 & 256 \\
Dropout                & 0 & 0 & 0  \\
In-context class tokens  & \multicolumn{3}{c}{32 (if used)} \\
In-context start block   & 4    & 8    & 10  \\
\hline
\multicolumn{3}{l}{\textcolor{gray}{Feature Pre-Processing}} \\
Pixels & \multicolumn{3}{c}{$[-1, 1]$ linear min-max rescaling} \\
\rowcolor{lightblue}
DINOv2 & \multicolumn{3}{c}{Patch-level scaling following RAE~\cite{zheng2025diffusion}} \\
\hline
\multicolumn{2}{l}{\textcolor{gray}{Training}} \\
Epochs                 & \multicolumn{3}{c}{200 (ablation), 600} \\
Warmup epochs~\cite{li2025back}  & \multicolumn{3}{c}{5} \\
Optimizer              & \multicolumn{3}{c}{Adam~\cite{kingma2014adam}, $\beta_1,\beta_2=0.9,0.95$} \\
Batch size             & \multicolumn{3}{c}{1024} \\
Learning rate          & \multicolumn{3}{c}{2e-4} \\
LR schedule            & \multicolumn{3}{c}{constant} \\
Weight decay           & \multicolumn{3}{c}{0} \\
EMA decay              & \multicolumn{3}{c}{$\{0.9996,\,0.9998,\,0.9999\}$} \\
Time sampler           & \multicolumn{3}{c}{$\text{logit}(t)\sim\mathcal{N}(\mu,\sigma^2),\ \mu=-0.8,\ \sigma=0.8$} \\
Noise scale            & \multicolumn{3}{c}{1.0} \\
Clip of $(1-t)$ in division  & \multicolumn{3}{c}{0.05} \\
\rowcolor{lightblue}
Class \& DINOv2 tokens joint dropout (for CFG)  & \multicolumn{3}{c}{0.1} \\
\rowcolor{lightblue}
Attention mask when dropout (for CFG) & \multicolumn{3}{c}{Semantic-to-pixel mask} \\
\rowcolor{lightblue}
$\lambda_d$ & \multicolumn{3}{c}{0.1} \\
\rowcolor{lightblue}
$\lambda_{\text{hyb}}$ & \multicolumn{3}{c}{10.0} \\
\rowcolor{lightblue}
$\tau_{\text{gate}}$ & \multicolumn{3}{c}{10.0} \\
\rowcolor{lightblue}
$\tau_{\text{rep}}$ & \multicolumn{3}{c}{0.2} \\
\hline

\multicolumn{2}{l}{\textcolor{gray}{Sampling}} \\
ODE solver             & \multicolumn{3}{c}{Heun~\cite{heun1900neue}} \\
ODE steps              & \multicolumn{3}{c}{50} \\
Time steps             & \multicolumn{3}{c}{linear in $[0.0,\,1.0]$} \\
CFG scale sweep range  & \multicolumn{3}{c}{$[1.0,\,4.0]$} \\
CFG interval           & \multicolumn{3}{c}{$[0.1,\,1]$ (if used)} \\
\hlineB{3}
\end{tabular}}
\end{center}
\end{table}

\begin{table}[t]
\begin{center}
\setlength{\tabcolsep}{0.65em}
\renewcommand{\arraystretch}{1.0}
\caption{\textbf{Comparison of architectural designs for visual co-denoising.} We compare baseline backbones, single-stream fusion strategies, and dual-stream fusion variants with different allocations of feature-specific and shared/dual-stream blocks. All variants keep the pixel stream depth fixed at 12 JiT blocks for fair comparison. JiT-B/16$^{\ddagger}$ and JiT-B/16$^{\dagger}$ denote widened variants with hidden dimensions increased from 768 to 1024 and 1088, respectively, to match the parameter counts of the dual-stream models. Blue rows mark the stronger variants used in subsequent analysis. Following previous works~\cite{li2025back, baade2026latent}, we mainly use FID as reference.
We highlight the default setting of the dual-stream model architecture in V-Co in \colorbox{lightblue}{light blue}.
}
\label{table:appendix_model_arch_ablation}
\scalebox{0.72}{
\begin{tabular}{clcccccc}
\hlineB{3}
 \multirow{2}{*}{} & \multirow{2}{*}{\textbf{Model}} & \multirow{2}{*}{\textbf{Backbone}} & \multirow{2}{*}{\textbf{\#Params}} & \textbf{\#Feature-Specific}  & \textbf{\#Shared/Dual-} & \multicolumn{2}{c}{\textbf{CFG=1.0}} \\
 \cline{7-8}
 & & & & \textbf{Blocks} & \textbf{Stream Blocks} & \textbf{FID$\downarrow$} & \textbf{IS$\uparrow$} \\ 
\hline
 \multicolumn{3}{l}{\textcolor{gray}{Baselines}} \\
(a) & JiT-B/16~\cite{li2025back}      & JiT-B/16            & 133M & - & -  & 32.54 & \textcolor{gray}{49.5}  \\
(b) &  JiT-B/16~\cite{li2025back}      & JiT-B/16$^{\dagger}$   & 261M & - & -  & 22.67 & \textcolor{gray}{69.9}  \\
(c) &  LatentForcing~\cite{baade2026latent} & JiT-B/16            & 156M & 2 & 10 & 13.06 & \textcolor{gray}{102.2}  \\
\hline
 \multicolumn{3}{l}{\textcolor{gray}{Single-Stream JiT Architecture}} \\
 (d) & DirectAddition& JiT-B/16               & 156M & 2 & 10 & 15.15 & \textcolor{gray}{103.4}  \\
 (e) & DirectAddition& JiT-B/16               & 177M & 4 & 8  & 12.90 & \textcolor{gray}{112.7}  \\
 (f) & DirectAddition& JiT-B/16               & 198M & 6 & 6  & 11.77 & \textcolor{gray}{116.2}  \\
 (g) & DirectAddition& JiT-B/16               & 220M & 8 & 4  & 14.20 & \textcolor{gray}{104.1}  \\
 (h) & DirectAddition& JiT-B/16               & 241M & 10& 2  & 14.43 & \textcolor{gray}{99.4}  \\
 (i) & ChannelConcat & JiT-B/16               & 157M & 2 & 10 & 14.33 & \textcolor{gray}{107.7}  \\
 (j) & ChannelConcat & JiT-B/16               & 178M & 4 & 8  & 11.93 & \textcolor{gray}{117.3}  \\
 (k) & ChannelConcat & JiT-B/16               & 200M & 6 & 6  & 11.23 & \textcolor{gray}{119.0}  \\
 (l) & ChannelConcat & JiT-B/16               & 221M & 8 & 4  & 13.73 & \textcolor{gray}{104.6}  \\
 (m) & ChannelConcat & JiT-B/16               & 242M & 10& 2  & 14.60 & \textcolor{gray}{99.7}  \\
(n)  & TokenConcat & JiT-B/16                 & 156M & 2 & 10 & 14.70 & \textcolor{gray}{103.8}   \\
(o)  &  TokenConcat & JiT-B/16                & 177M & 4 & 8  & 12.59 & \textcolor{gray}{112.8}   \\
(p)  & TokenConcat  & JiT-B/16                & 198M & 6 & 6  & 12.35 & \textcolor{gray}{116.7}   \\
(q)  & TokenConcat  & JiT-B/16                & 220M & 8 & 4  & 14.31 &  \textcolor{gray}{104.4} \\
(r)  & TokenConcat  & JiT-B/16                & 241M & 10& 2  & 14.97 &  \textcolor{gray}{99.4}  \\
(s)  & TokenConcat  & JiT-B/16$^{\ddagger}$   & 265M & 6 & 6  & 9.74  &  \textcolor{gray}{129.4} \\
(t)  & TokenConcat  & JiT-B/16$^{\ddagger}$   & 274M & 8 & 4  & 11.43 &  \textcolor{gray}{118.1} \\
(u)  & TokenConcat  & JiT-B/16$^{\ddagger}$   & 284M & 10& 2  & 12.90 &  \textcolor{gray}{109.4} \\
\hline
\multicolumn{3}{l}{\textcolor{gray}{Dual-Stream JiT Architecture}}  \\
\rowcolor{lightblue}
(v) & TokenConcat & JiT-B/16            & 260M & 6 & 6  & 11.78 & {\textcolor{gray}{115.4}} \\
\rowcolor{lightblue}
(w) &  TokenConcat & JiT-B/16            & 260M & 4 & 8  & 11.40 & \textcolor{gray}{118.3} \\
\rowcolor{lightblue}
(x) & TokenConcat & JiT-B/16            & 260M & 2 & 10 & 10.24 & \textcolor{gray}{124.5} \\
\rowcolor{lightblue}
(y) & TokenConcat & JiT-B/16            & 260M & 0 & 12 & \textbf{8.86} & \textcolor{gray}{\textbf{132.8}}  \\
\hlineB{3}
\end{tabular}} 
\end{center}
\end{table}

Specifically, in Sec. 3.2, we start from a minimal co-denoising setup. At this stage, the only additional component is the DINOv2~\cite{oquab2023dinov2} branch in the feature preprocessing stage, where DINOv2-Base features are normalized using the dataset-level statistics computed in RAE~\cite{zheng2025diffusion}. For conditioning dropout, we adopt the standard independent dropout strategy, with class-label dropout set to 0.1 and DINOv2 feature dropout set to 0.2, applied separately rather than jointly. No attention mask is used in this section. The DINOv2 denoising loss coefficient (\ie{}, $\lambda_d$) is set to 0.1 from this stage onward.
Starting from Sec. 3.3, we further refine the conditioning design. As shown in Table 2 and~\Cref{table:label_dino_dropout}, jointly dropping class labels and DINOv2 features yields the best performance, and we therefore adopt a joint dropout probability of 0.1 from this section onward. Moreover, the semantic-to-pixel attention mask achieves the strongest results in Table 2, and is thus used as the default setting in the subsequent experiments.
In Sec. 3.4, we further introduce the perceptual-drifting hybrid loss to improve feature alignment during co-denoising. The corresponding hyper-parameters, including $\lambda_{\text{hyb}}$, $\tau_{\text{gate}}$, and $\tau_{\text{rep}}$, are introduced from this section onward.
Finally, in Sec. 3.5, we compare noise-schedule shifting and RMS scaling for feature calibration. Since RMS scaling performs best in our ablation, we adopt it as the default calibration strategy in the final model.

\begin{table}[t]
\begin{center}
\setlength{\tabcolsep}{0.65em}
\renewcommand{\arraystretch}{1.0}
\caption{\textbf{Comparison of different label/DINO dropout strategies under unguided (\ie{}, CFG=1.0) and guided (\ie{}, CFG=2.9) generation.} The best and second best numbers are \textbf{bolded} and \underline{underlined}.
We highlight the default setting of joint dropout of class labels and DINOv2 features in V-Co in \colorbox{lightblue}{light blue}.
}
\label{table:label_dino_dropout}
\scalebox{0.77}{
\begin{tabular}{lllcccc}
\hlineB{3}
 \multirow{2}{*}{Label Dropout Ratio} & \multirow{2}{*}{DINO Dropout Ratio} 
& \multicolumn{2}{c}{CFG=1.0} & \multicolumn{2}{c}{CFG=2.9} \\
\cmidrule(lr){3-4} \cmidrule(lr){5-6}
 & & FID$\downarrow$ & IS$\uparrow$ & FID$\downarrow$ & IS$\uparrow$ \\
\hline
 \multicolumn{2}{l}{\textcolor{gray}{Independent Dropout}} \\
 0.1 & 0.1 & 7.01 & 136.37 & 3.77 & 189.38 \\
0.1 & 0.2 & 7.28 & 136.84 & {3.59} & 189.69 \\
0.1 & 0.3 & 7.78 & 129.55 & 3.73 & 188.38 \\
0.2 & 0.1 & 7.50 & 128.84 & 4.11 & 175.04 \\
0.2 & 0.2 & 7.47 & 131.44 & 3.98 & 180.24 \\
0.2 & 0.3 & 9.04 & 117.94 & 4.11 & 173.21 \\
0.3 & 0.1 & 8.80 & 117.98 & 4.38 & 165.06 \\
0.3 & 0.2 & 8.40 & 122.23 & 4.09 & 172.15 \\
0.3 & 0.3 & 9.57 & 114.64 & 4.52 & 165.55 \\
\hline
 \multicolumn{2}{l}{\textcolor{gray}{Joint Dropout}} \\
\rowcolor{lightblue}
{0.1} & {0.1} & \underline{5.38} & \textbf{161.4} & 3.55 & 214.39 \\
\rowcolor{lightblue}
0.2 & 0.2 & 5.62 & 158.51 & \underline{3.18} & \textbf{219.60} \\
\rowcolor{lightblue}
0.3 & 0.3 & \textbf{5.17} & \underline{159.92} & \textbf{3.17} & \underline{219.41} \\
\hlineB{3}
\end{tabular}}
\end{center}
\end{table}

\section{Additional Ablations}
\label{subsec:appendix_additional_ablations}

\noindent\textbf{Extended ablation of single-stream fusion strategies.}
In~\Cref{table:appendix_model_arch_ablation}, we present additional quantitative results that complement Table 1 in the main paper. Specifically, we study different allocations of feature-specific and shared blocks under three interaction strategies: direct addition, channel concatenation, and token concatenation. Overall, we observe that a balanced design, with 6 feature-specific blocks and 6 shared blocks out of 12 total blocks, generally yields the best performance. For the token-concatenation strategy, we further examine widened variants by increasing the hidden dimension from 768 to 1024, resulting in models with 265M to 284M parameters. Nevertheless, none of these variants surpasses our default dual-stream co-denoising design, which achieves the best performance with 260M parameters.

\noindent\textbf{Extended ablation of label and DINOv2 dropout strategies.}
In~\Cref{table:label_dino_dropout}, we provide additional quantitative results complementing Table 2 in the main paper. Specifically, we evaluate different dropout ratios for labels and DINOv2 features, applied either independently or jointly. The results indicate that independently dropping labels or DINOv2 features generally under-performs joint dropout.  
As discussed in the main paper, once the unconditional prediction is structurally defined through semantic-to-pixel masking, removing all conditioning inputs from the pixel branch during training leads to better alignment with inference-time behavior.

\begin{table}[t]
\begin{center}
\setlength{\tabcolsep}{0.65em}
\renewcommand{\arraystretch}{1.0}
\caption{\textbf{Ablation of the similarity-based gating $s_i$ in Eq. (13).}
We conduct an ablation using a simplified version of $s_i$, where the gate is replaced with a scalar value $s$ instead of being dependent on real and generated samples. 
We report both unguided and guided FID and IS scores while sweeping the scalar $s$. Row with the best guided FID score is highlighted in \colorbox{lightblue}{light blue}.
}
\label{table:appendix_linear_gate_tau_ablation}
\scalebox{0.85}{
\begin{tabular}{lcccc}
\hlineB{3}
\multirow{2}{*}{\textbf{$s$}}  & \multicolumn{2}{c}{\textbf{Unguided (CFG=1.0)}} & \multicolumn{2}{c}{\textbf{Guided (Best CFG$>1.0$)}}  \\  \cline{2-3} \cline{4-5}  
 &  \textbf{FID$\downarrow$} & \textbf{IS$\uparrow$} & \textbf{FID$\downarrow$} & \textbf{IS$\uparrow$} \\  
\hline
 \multicolumn{2}{l}{\textcolor{gray}{Scalar gate $s$}} \\
  0     &  {15.72}       & \textcolor{gray}{{80.1}}           & {8.13}        & \textcolor{gray}{{114.4}} \\
  0.001 &  {11.24}       & \textcolor{gray}{{114.7}}          & {4.24}        & \textcolor{gray}{{171.7}} \\
  0.01  &  {11.81}       & \textcolor{gray}{{109.9}}          & {4.46}        & \textcolor{gray}{{165.6}} \\
  0.1   &  {13.22}       & \textcolor{gray}{{99.2}}           & {5.35}        & \textcolor{gray}{{151.6}} \\
  0.5   & {10.40} & \textcolor{gray}{\textbf{186.0}}   & {5.68} & \textcolor{gray}{\textbf{253.3}} \\
  0.9   &  {5.48}        & \textcolor{gray}{{174.7}}          & {2.75}        & \textcolor{gray}{{233.9}} \\
  0.99  &  \textbf{5.33}        & \textcolor{gray}{{183.4}}          & {2.83}        & \textcolor{gray}{{246.7}} \\
  0.999 &  {5.41}        & \textcolor{gray}{{182.6}}          & {2.76}        & \textcolor{gray}{{249.8}} \\
  \hline 
 \multicolumn{4}{l}{\textcolor{gray}{Similarity-Based Gating $s_i$ (Default in V-Co)}} \\
 \rowcolor{lightblue}
  $s_i$ &  {5.17}        & \textcolor{gray}{{181.6}}          & \textbf{2.61}        & \textcolor{gray}{{243.9}} \\
\hlineB{3}
\end{tabular}}
\end{center}
\end{table}

\begin{table}[t]
\begin{center}
\setlength{\tabcolsep}{0.65em}
\renewcommand{\arraystretch}{1.0}
\caption{\textbf{Comparison of different repulsion temperatures $\tau_{\text{rep}}$.} We report both unguided and guided FID and IS scores by sweeping the repulsion temperature $\tau_{\text{rep}}$. We highlight the rows with the best guided FID scores in \colorbox{lightblue}{light blue}.
}
\label{table:appendix_repulsion_tau_ablation}
\scalebox{0.8}{
\begin{tabular}{lcccc}
\hlineB{3}
\multirow{2}{*}{\textbf{$\tau_{\text{rep}}$}}  & \multicolumn{2}{c}{\textbf{Unguided (CFG=1.0)}} & \multicolumn{2}{c}{\textbf{Guided (Best CFG$>1.0$)}}  \\  \cline{2-3} \cline{4-5}  
 &  \textbf{FID$\downarrow$} & \textbf{IS$\uparrow$} & \textbf{FID$\downarrow$} & \textbf{IS$\uparrow$} \\  
\hline
  2e-3 &  {4.97} & \textcolor{gray}{{186.4}}          & {2.72} & \textcolor{gray}{{251.3}} \\
  2e-2 &  \textbf{4.92} & \textcolor{gray}{\textbf{186.7}}          & {2.66} & \textcolor{gray}{\textbf{251.7}} \\
   \rowcolor{lightblue}
  2e-1 &  {5.17} & \textcolor{gray}{{181.6}}          & \textbf{2.61} & \textcolor{gray}{{243.9}} \\
  2 &  {5.33} & \textcolor{gray}{{180.1}}          & {2.64} & \textcolor{gray}{{244.8}} \\
  2e1 &  {5.30} & \textcolor{gray}{{180.7}}          & {2.67} & \textcolor{gray}{{247.7}} \\
\hlineB{3}
\end{tabular}}
\end{center}
\end{table}

\begin{table}[t!]
\begin{center}
\setlength{\tabcolsep}{0.65em}
\renewcommand{\arraystretch}{1.0}
\caption{\textbf{Comparison of different gate temperatures $\tau_{\text{gate}}$.} We report both unguided and guided FID and IS scores by sweeping the gate temperature $\tau_{\text{gate}}$. We highlight the rows with the best guided FID scores in \colorbox{lightblue}{light blue}.
}
\label{table:appendix_gate_tau_ablation}
\scalebox{0.8}{
\begin{tabular}{lcccc}
\hlineB{3}
\multirow{2}{*}{\textbf{$\tau_{\text{gate}}$}}  & \multicolumn{2}{c}{\textbf{Unguided (CFG=1.0)}} & \multicolumn{2}{c}{\textbf{Guided (Best CFG$>1.0$)}}  \\  \cline{2-3} \cline{4-5}  
 &  \textbf{FID$\downarrow$} & \textbf{IS$\uparrow$} & \textbf{FID$\downarrow$} & \textbf{IS$\uparrow$} \\  
\hline
  1e-2 &  {11.66} & \textcolor{gray}{{112.2}}          & {4.01} & \textcolor{gray}{{173.5}} \\
  1e-1 &  {10.96} & \textcolor{gray}{{114.6}}          & {4.06} & \textcolor{gray}{{173.6}} \\
  1 &  {7.34} & \textcolor{gray}{{136.9}}          & {3.10} & \textcolor{gray}{{189.6}} \\
  1e1 &  {5.41} & \textcolor{gray}{{108.8}}          & {2.75} & \textcolor{gray}{{247.7}} \\
  \rowcolor{lightblue}
  1e2 &  {5.17} & \textcolor{gray}{{181.6}}          & \textbf{2.61} & \textcolor{gray}{{243.9}} \\
  1e3 &  \textbf{5.11} & \textcolor{gray}{\textbf{185.8}}          & {2.77} & \textcolor{gray}{\textbf{252.2}} \\
\hlineB{3}
\end{tabular}}
\end{center}
\end{table}

\noindent\textbf{Ablation of the similarity-based gating $s_i$ in Eq. (13).}
To evaluate the effectiveness of the proposed similarity-based gating mechanism, we conduct an ablation where the adaptive gate $s_i$ is replaced with a scalar value $s$, removing its dependence on real and generated samples. Under this simplification, the hybrid potential becomes
$V_{\mathrm{hyb}}(\bm{u}_i) = s \cdot V_{\mathrm{pos}}(\bm{u}_i) - (1 - s) \cdot V_{\mathrm{neg}}(\bm{u}_i)$.
We report both unguided and guided FID and IS scores while sweeping the scalar $s$. As shown in~\Cref{table:appendix_linear_gate_tau_ablation}, the default similarity-based gating in V-Co achieves the best guided FID score compared with the simplified scalar gate $s$, demonstrating the effectiveness of our design.

\begin{table}[t]
\begin{center}
\setlength{\tabcolsep}{0.65em}
\renewcommand{\arraystretch}{1.0}
\caption{\textbf{Comparison of different hybrid loss coefficients $\lambda_{\text{hyb}}$.} We report both unguided and guided FID and IS scores by sweeping the hybrid loss coefficients $\lambda_{\text{hyb}}$. We highlight the rows with the best guided FID scores in \colorbox{lightblue}{light blue}.
}
\label{table:appendix_hybrid_loss_coefficients}
\scalebox{0.8}{
\begin{tabular}{lcccc}
\hlineB{3}
\multirow{2}{*}{\textbf{$\lambda_{\text{hyb}}$}}  & \multicolumn{2}{c}{\textbf{Unguided (CFG=1.0)}} & \multicolumn{2}{c}{\textbf{Guided (Best CFG$>1.0$)}}  \\  \cline{2-3} \cline{4-5}  
 &  \textbf{FID$\downarrow$} & \textbf{IS$\uparrow$} & \textbf{FID$\downarrow$} & \textbf{IS$\uparrow$} \\  
\hline
  1e-2 &  {7.10} & \textcolor{gray}{{139.6}}          & {3.30} & \textcolor{gray}{{187.0}} \\
  1e-1 &  {6.72} & \textcolor{gray}{{142.5}}          & {3.17} & \textcolor{gray}{{189.9}} \\
  1 &  {5.23} & \textcolor{gray}{{162.5}}          & {2.74} & \textcolor{gray}{{214.0}} \\
  \rowcolor{lightblue}
  1e1 &  \textbf{5.17} & \textcolor{gray}{\textbf{181.6}}          & \textbf{2.61} & \textcolor{gray}{\textbf{243.9}} \\
  1e2 &  {13.61} & \textcolor{gray}{{133.4}}          & {4.49} & \textcolor{gray}{{221.4}} \\
  1e3 &  {31.30} & \textcolor{gray}{{79.3}}          & {17.90} & \textcolor{gray}{{124.4}} \\
  1e4 &  {65.84} & \textcolor{gray}{{41.8}}          & {58.04} & \textcolor{gray}{{56.0}} \\
\hlineB{3}
\end{tabular}}
\end{center}
\end{table}

\begin{table}[t!]
\begin{center}
\setlength{\tabcolsep}{0.65em}
\renewcommand{\arraystretch}{1.0}
\caption{\textbf{Comparison of different DINOv2 model sizes.} We report both unguided and guided FID and IS scores while sweeping the DINOv2 diffusion loss coefficient $\lambda_d$ over $\{1\mathrm{e}{-3}, 1\mathrm{e}{-2}, 1\mathrm{e}{-1}, 1\}$. We highlight the rows with the best guided FID scores in \colorbox{lightblue}{light blue} for each DINOv2 model size.
}
\label{table:appendix_dinov2_size_ablations}
\scalebox{0.85}{
\begin{tabular}{lcccccc}
\hlineB{3}
\multirow{2}{*}{\textbf{Model}} & \multirow{2}{*}{\textbf{$\lambda_d$}} & \multirow{2}{*}{\textbf{\#Params}} & \multicolumn{2}{c}{\textbf{Unguided (CFG=1.0)}} & \multicolumn{2}{c}{\textbf{Guided (Best CFG$>1.0$)}}  \\  \cline{4-5} \cline{6-7}  
 & &  & \textbf{FID$\downarrow$} & \textbf{IS$\uparrow$} & \textbf{FID$\downarrow$} & \textbf{IS$\uparrow$} \\  
\hline
 DINOv2-Small & 1e-3 & 22M  & {9.04} & \textcolor{gray}{{118.2}} & {5.06} & \textcolor{gray}{{156.4}} \\
  \rowcolor{lightblue}
 DINOv2-Small & 1e-2 & 22M  & {6.70} & \textcolor{gray}{{134.4}} & \textbf{3.67} & \textcolor{gray}{\textbf{176.2}} \\
 DINOv2-Small & 1e-1 & 22M  & \textbf{6.35} & \textcolor{gray}{\textbf{140.0}} & {4.11} & \textcolor{gray}{{174.8}} \\
 DINOv2-Small & 1    & 22M  & {9.05} & \textcolor{gray}{{126.0}} & {7.97} & \textcolor{gray}{{145.3}} \\
\hline
 DINOv2-Base  & 1e-3 & 86M  & {12.27}        & \textcolor{gray}{{107.7}}        & {6.45} & \textcolor{gray}{{151.9}} \\
 DINOv2-Base  & 1e-2 & 86M  & {9.81}         & \textcolor{gray}{{118.2}}        & {5.16} & \textcolor{gray}{{163.2}} \\
\rowcolor{lightblue}
 DINOv2-Base  & 1e-1 & 86M  & \textbf{6.35}  & \textcolor{gray}{\textbf{140.0}} & \textbf{4.11} & \textcolor{gray}{\textbf{174.8}} \\
 DINOv2-Base  & 1    & 86M  & 16.77          & \textcolor{gray}{95.1}           & 16.48          & \textcolor{gray}{106.9} \\
\hline
 DINOv2-Large & 1e-3 & 304M & 13.92         & \textcolor{gray}{96.3}           & 6.59          & \textcolor{gray}{143.1} \\
 \rowcolor{lightblue}
 DINOv2-Large & 1e-2 & 304M & 9.19          & \textcolor{gray}{119.9}          & \textbf{4.20} & \textcolor{gray}{173.8} \\
 DINOv2-Large & 1e-1 & 304M & \textbf{8.70} & \textcolor{gray}{\textbf{124.3}} & 4.57          & \textcolor{gray}{\textbf{174.7}} \\
 DINOv2-Large & 1    & 304M & 32.28         & \textcolor{gray}{82.2}           & 25.32         & \textcolor{gray}{94.0}  \\
\hline
 DINOv2-Giant & 1e-3 & 1.1B & 13.15         & \textcolor{gray}{99.3}           & 7.46          & \textcolor{gray}{143.4} \\
 DINOv2-Giant & 1e-2 & 1.1B & 10.41         & \textcolor{gray}{112.1}          & 5.42          & \textcolor{gray}{160.5} \\
  \rowcolor{lightblue}
 DINOv2-Giant & 1e-1 & 1.1B & \textbf{8.91} &  \textcolor{gray}{\textbf{120.1}} & \textbf{5.00} & \textcolor{gray}{\textbf{166.7}} \\
 DINOv2-Giant & 1    & 1.1B & 23.02         & \textcolor{gray}{83.4}           & 24.18         & \textcolor{gray}{95.9}  \\
\hlineB{3}
\end{tabular}}
\end{center}
\end{table}

\noindent\textbf{Hyper-parameter tuning for the perceptual-drifting hybrid loss.}
In~\Cref{table:appendix_repulsion_tau_ablation,table:appendix_gate_tau_ablation,table:appendix_hybrid_loss_coefficients}, we perform hyper-parameter sweeps over the three key hyper-parameters in the perceptual-drifting hybrid loss: the repulsion temperature $\tau_{\text{rep}}$, the gate temperature $\tau_{\text{gate}}$, and the hybrid loss coefficient $\lambda_{\text{hyb}}$. The default hyper-parameters used in V-Co are selected from these settings based on the best guided FID scores.

\noindent\textbf{Comparison of different DINOv2 model sizes.}
In~\Cref{table:appendix_dinov2_size_ablations}, we compare V-Co trained with DINOv2 features of different model sizes as semantic representations for co-denoising with pixels. For each DINOv2 model, we re-compute the feature scaling factor based on its RMS value to ensure that the SNR ratio between the DINOv2 features and pixels remains consistent. We also sweep over different DINOv2 diffusion loss coefficients $\lambda_d$, as different encoder sizes may perform best under different loss scales. The results show that even relatively small representation encoders preserve sufficient low-level detail for co-denoising. Similar experiment findings trend also observed in RAE~\cite{zheng2025diffusion} (\ie{}, Table 1(c)).

\TODO{\noindent\textbf{Generalization to other visual encoders.}
Although V-Co uses DINOv2 as the default semantic encoder, our recipe is not tied to a specific representation model. 
To verify this, we replace DINOv2 with three different pretrained visual encoders: CLIP-B/16, SigLIP2-L/16, and MAE-L/16. 
For each encoder, we compare the full V-Co recipe against variants that remove the perceptual-drifting hybrid loss and/or RMS-based feature scaling. 
As shown in Table~\ref{table:other_encoders}, the full recipe consistently achieves the best FID and IS across all three encoders. 
This suggests that the two key components introduced in V-Co, RMS-based calibration and the perceptual-drifting hybrid loss, provide general benefits beyond DINOv2 features.}

\begin{table}[t]
\begin{center}
\setlength{\tabcolsep}{0.55em}
\renewcommand{\arraystretch}{1.0}
\caption{\textbf{Ablation with different pretrained visual encoders.}
We replace the default DINOv2 encoder with MAE-L/16, SigLIP2-L/16, and CLIP-B/16. 
For each encoder, we compare the full V-Co recipe against variants without the perceptual-drifting hybrid loss and/or RMS-based feature scaling. 
The full recipe achieves the best FID and IS across all encoders, showing that the effectiveness of V-Co is not limited to DINOv2 representations.}
\label{table:other_encoders}
\scalebox{0.90}{
\begin{tabular}{lcccccc}
\hlineB{3}
\multirow{2}{*}{\textbf{Encoder}}
& \multicolumn{2}{c}{\textbf{RMS + Hybrid}}
& \multicolumn{2}{c}{\textbf{RMS Only}}
& \multicolumn{2}{c}{\textbf{w/o RMS or Hybrid}} \\
\cline{2-7}
& \textbf{IS$\uparrow$} & \textbf{FID$\downarrow$}
& \textbf{IS$\uparrow$} & \textbf{FID$\downarrow$}
& \textbf{IS$\uparrow$} & \textbf{FID$\downarrow$} \\
\hline
MAE-L/16
& \textbf{139.90} & \textbf{9.63}
& 126.46 & 11.18
& 93.00 & 17.92 \\
SigLIP2-L/16
& \textbf{229.10} & \textbf{2.79}
& 67.50 & 19.34
& 186.43 & 3.52 \\
CLIP-B/16
& \textbf{237.43} & \textbf{2.84}
& 202.50 & 3.06
& 173.00 & 3.86 \\
\hlineB{3}
\end{tabular}}
\end{center}
\end{table}

\TODO{\noindent\textbf{Higher-resolution ImageNet generation.}
To further evaluate whether V-Co remains effective beyond the ImageNet-$256$ setting used in most ablations, we train V-Co-B/16 on ImageNet-$512$ following the $512{\times}512$ setup of JiT~\cite{li2025back}. 
Table~\ref{table:higher_resolution} compares V-Co-B/16 with JiT-B/16 and JiT-L/16 at both $200$ and $600$ training epochs. 
On ImageNet-$512$, V-Co-B/16 achieves FID $2.93$ after $200$ epochs and FID $2.42$ after $600$ epochs, outperforming both JiT-B/16 and JiT-L/16 under the same training lengths. 
This trend is consistent with the ImageNet-$256$ results, indicating that the V-Co recipe remains effective when scaling to higher image resolution.}

\begin{table}[t]
\begin{center}
\setlength{\tabcolsep}{0.80em}
\renewcommand{\arraystretch}{1.0}
\caption{\textbf{Higher-resolution ImageNet results.}
We compare FID at $200$ and $600$ training epochs on ImageNet-$512$ and ImageNet-$256$. 
V-Co-B/16 consistently outperforms JiT-B/16 and JiT-L/16 at both resolutions, showing that the proposed co-denoising recipe remains effective at higher resolution.}
\label{table:higher_resolution}
\scalebox{0.90}{
\begin{tabular}{lccc}
\hlineB{3}
\textbf{FID @ 200 / 600 Epochs$\downarrow$}
& \textbf{JiT-B/16}
& \textbf{JiT-L/16}
& \textbf{V-Co-B/16} \\
\hline
ImageNet-\textbf{512}
& 4.64 / 4.02
& 3.06 / 2.53
& \cellcolor{lightblue}\textbf{2.93} / \textbf{2.42} \\
ImageNet-\textbf{256}
& 4.37 / 3.66
& 2.79 / 2.36
& \cellcolor{lightblue}\textbf{2.70} / \textbf{2.35} \\
\hlineB{3}
\end{tabular}}
\end{center}
\end{table}

\section{Generated Samples}
\label{subsec:appendix_generated_samples}
\begin{figure*}[t]
    \centering
    \includegraphics[width=0.99\linewidth]{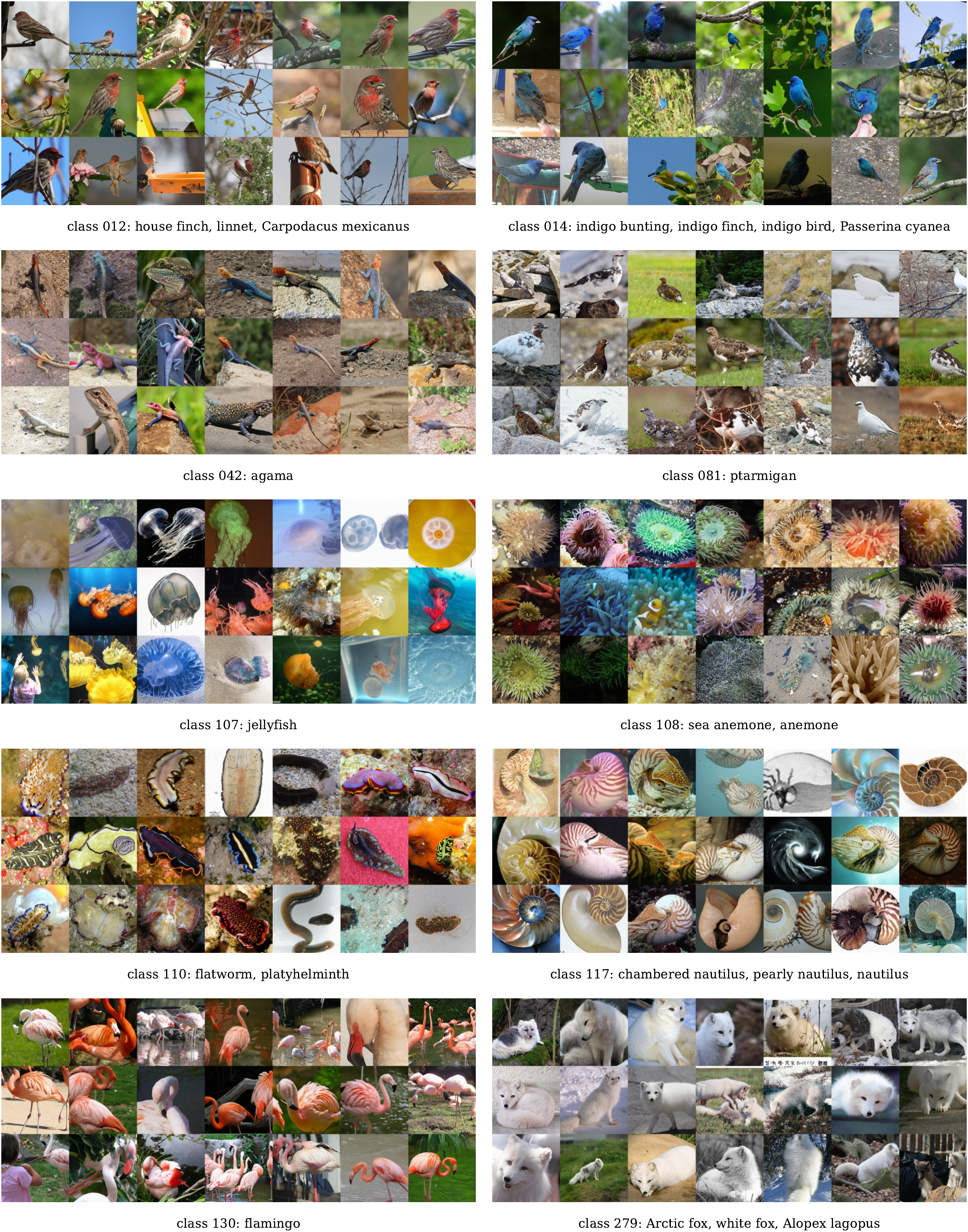}
    \caption{\emph{Uncurated}
    samples on ImageNet 256$\times$256 using V-Co-H/16 conditioned on the specified classes. Unlike the common practice of visualizing with a higher CFG, here we show images using the CFG value (1.5) that achieves the reported FID of 1.71.}   
\label{fig:imagenet_samples_1}
\end{figure*}

\begin{figure*}[t]
    \centering
    \includegraphics[width=0.99\linewidth]{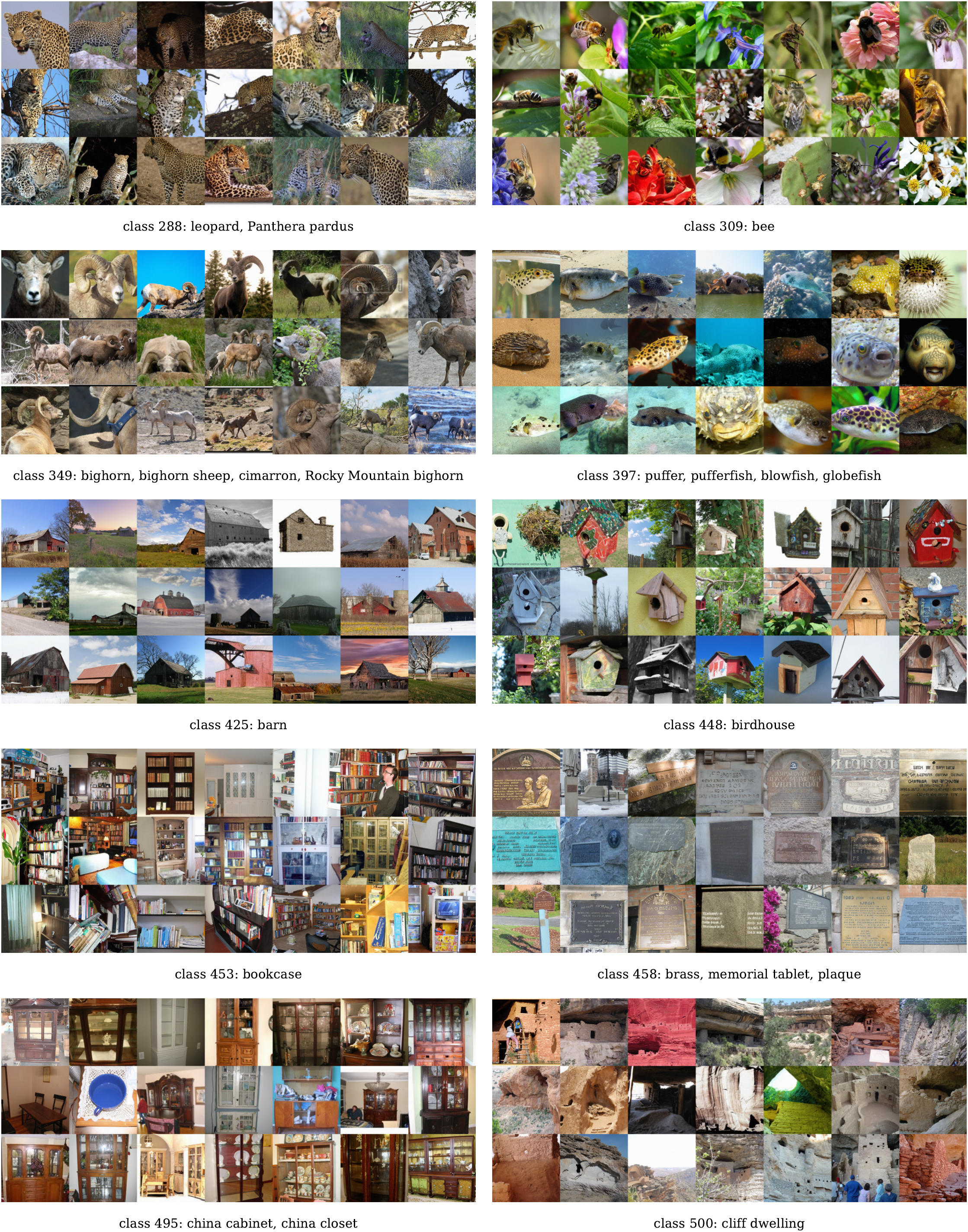}
    \caption{\emph{Uncurated}
    samples on ImageNet 256$\times$256 using V-Co-H/16 conditioned on the specified classes. Unlike the common practice of visualizing with a higher CFG, here we show images using the CFG value (1.5) that achieves the reported FID of 1.71.}   
\label{fig:imagenet_samples_2}
\end{figure*}

\begin{figure*}[t]
    \centering
    \includegraphics[width=0.99\linewidth]{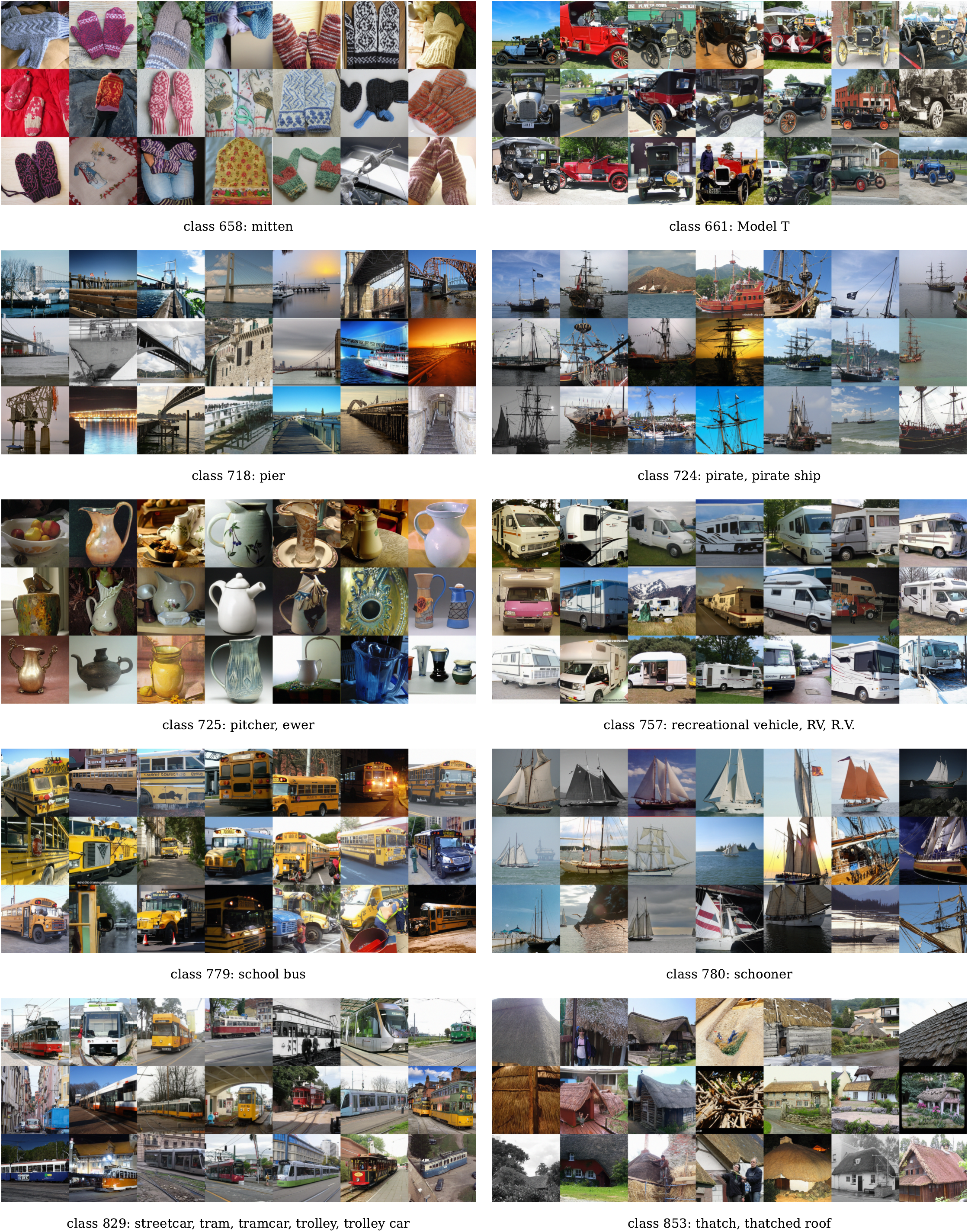}
    \caption{\emph{Uncurated}
    samples on ImageNet 256$\times$256 using V-Co-H/16 conditioned on the specified classes. Unlike the common practice of visualizing with a higher CFG, here we show images using the CFG value (1.5) that achieves the reported FID of 1.71.}   
\label{fig:imagenet_samples_3}
\end{figure*}

\begin{figure*}[t]
    \centering
    \includegraphics[width=0.99\linewidth]{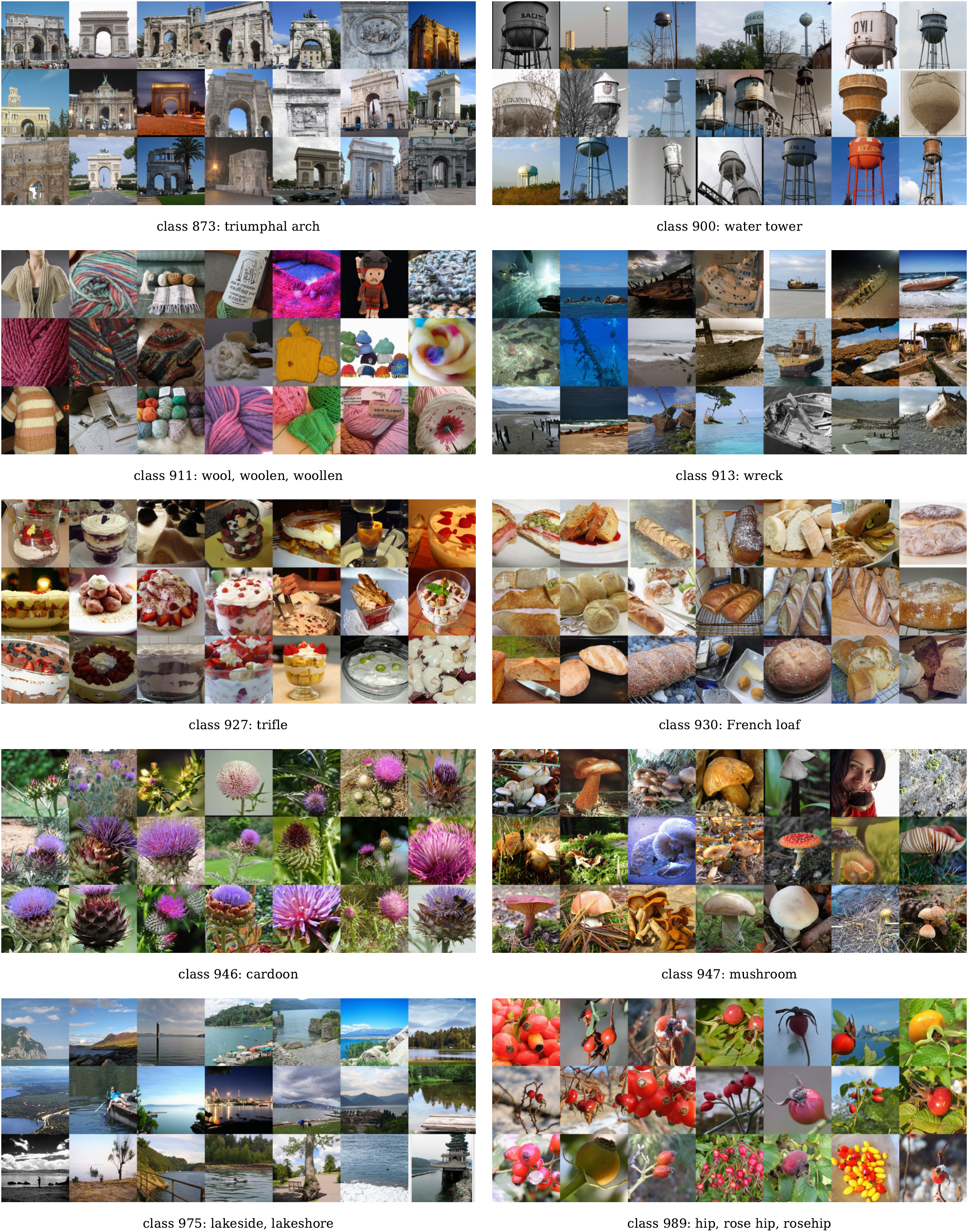}
    \caption{\emph{Uncurated}
    samples on ImageNet 256$\times$256 using V-Co-H/16 conditioned on the specified classes. Unlike the common practice of visualizing with a higher CFG, here we show images using the CFG value (1.5) that achieves the reported FID of 1.71.}   
\label{fig:imagenet_samples_4}
\end{figure*}

In~\cref{fig:imagenet_samples_1,fig:imagenet_samples_2,fig:imagenet_samples_3,fig:imagenet_samples_4}, we present \emph{uncurated} ImageNet 256$\times$256 samples generated by V-Co-H/16 after 300 epochs of training, conditioned on the specified classes. Unlike the common practice of using a larger CFG value for visualization, we instead show samples generated with the same CFG value (1.5) used to obtain the reported FID of 1.71.

\section{Reproducibility}
\label{subsec:reproduce}
Our complete training and inference code is included in the supplementary material to ensure reproducibility.

\section{Limitation and Future Work}
\label{sec:appendix_limitations}

While V-Co provides a clear and effective recipe for visual co-denoising in pixel-space diffusion, several limitations remain. First, our study focuses on class-conditional generation on ImageNet-256, which offers a controlled setting for isolating the effects of architecture, CFG design, auxiliary objectives, and feature calibration, but does not capture the full diversity of open-ended generation settings such as text-to-image synthesis or more structured multimodal tasks. Extending the proposed recipe beyond ImageNet-style class conditioning is therefore an important direction for future work.

Second, V-Co relies on pretrained semantic features from a strong external visual encoder (\ie{}, DINOv2). While this design is well aligned with our representation-alignment perspective and substantially improves semantic supervision in pixel-space generation, the resulting co-denoising dynamics may still depend on the quality, inductive biases, and spatial granularity of the teacher representation. Exploring alternative semantic feature sources is another promising direction.

Finally, our method is intentionally minimalist and does not incorporate stronger auxiliary supervision, such as combining REPA-style objectives with our perceptual-drifting hybrid loss. This keeps the empirical conclusions clean, but also suggests that the current system may not yet fully realize the potential of co-denoising. Future work may further explore how the V-Co recipe interacts with richer objectives and stronger supervision.

